\documentclass{article} 
\usepackage{iclr2026_conference,times}

\input{math_commands.tex}

\usepackage{hyperref}
\usepackage{url}
\usepackage{graphicx}
\usepackage{subcaption}
\usepackage{enumitem}
\usepackage{xspace}
\usepackage{amsmath}
\usepackage{cleveref}
\usepackage{booktabs}
\usepackage{multirow}
\usepackage{array}
\usepackage{colortbl}
\usepackage{wrapfig}
\usepackage[table]{xcolor}

\usepackage[most]{tcolorbox}
\usepackage{listings}
\usepackage{xcolor}
\usepackage{etoolbox} 
\newtcolorbox{prompt}{
  colback=black!2,    
  colframe=black!15,  
  boxrule=0.4pt,
  sharp corners,
  left=8pt,right=8pt,top=6pt,bottom=6pt,
  before skip=6pt, after skip=6pt,
  breakable,                 
  before upper=\ttfamily\small 
}

\definecolor{RubricNavy}{RGB}{16,36,132}    
\definecolor{RubricBack}{RGB}{235,238,249}  
\definecolor{RubricFrame}{RGB}{16,36,132}   

\newtcolorbox{rubricbox}[1]{%
  enhanced, breakable,
  colback=RubricBack,
  colframe=RubricFrame,
  colbacktitle=RubricNavy, coltitle=white,
  title=\sffamily\bfseries #1,
  boxrule=0.8pt, arc=2.5mm,
  left=8pt,right=8pt,top=8pt,bottom=8pt,
  before skip=8pt, after skip=8pt,
  before upper=\ttfamily\small 
                 \setlength{\parindent}{0pt}%
               \setlength{\parskip}{4pt}%

}

\crefname{section}{\S}{\S\S}
\crefname{figure}{fig.}{figs.}%
\crefname{appendix}{app.}{app.}%

\newtoggle{isiclr}
\toggletrue{isiclr}

\iftoggle{isiclr}{
    \newcommand{\ourmodelsmall}{{FARE-8B}\xspace} 
    \newcommand{\ourmodellarge}{{FARE-20B}\xspace} 
    \newcommand{\ourmodel}{{FARE}\xspace} 
}{
    \newcommand{\ourmodelsmall}{{SFR-FRJ-8B}\xspace} 
    \newcommand{\ourmodellarge}{{SFR-FRJ-20B}\xspace} 
    \newcommand{\ourmodel}{{SFR-FRJs}\xspace} 
}

\newtoggle{comments}
\toggletrue{comments}
\togglefalse{comments}

\iftoggle{comments}{
    \newcommand{\yilun}[1]{\textcolor{blue}{(Yilun: #1)}}
    \newcommand{\austin}[1]{\textcolor{orange}{(Austin: #1)}}
    \newcommand{\jason}[1]{\textcolor{green}{(Jason: #1)}}
    \newcommand{\shafiq}[1]{\textcolor{cyan}{(shafiq: #1)}}
    \newcommand{\nxphi}[1]{\textcolor{red}{(Phi: #1)}}
}{
    \newcommand{\yilun}[1]{}
    \newcommand{\austin}[1]{}
    \newcommand{\jason}[1]{}
    \newcommand{\shafiq}[1]{}
    \newcommand{\nxphi}[1]{}
}

\title{Foundational Automatic Evaluators:\\Scaling Multi-Task Generative Evaluator Training for Reasoning-Centric Domains}

\newcommand{\symbolimg}[2][0.3cm]{%
  \ensuremath{\vcenter{\hbox{\includegraphics[height=#1]{#2}}}}%
}

\newcommand{\huggingface}{\symbolimg[0.3cm]{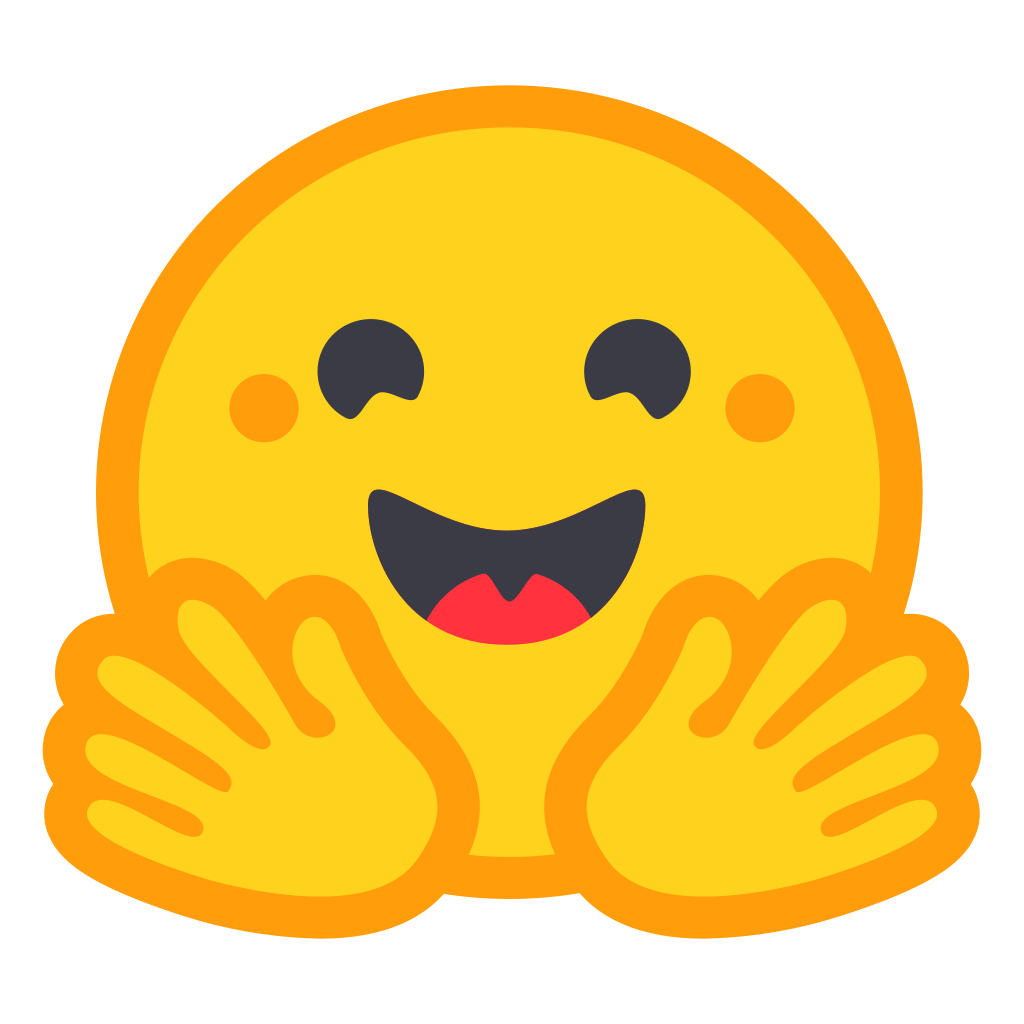}}

\author{
\parbox{\textwidth}{
{
    \centering
    \normalfont
    \vspace{1em} 
    \textbf{Austin Xu}, \textbf{Xuan-Phi Nguyen}, \textbf{Yilun Zhou}\thanks{Work done at Salesforce AI Research}, \textbf{Chien-Sheng Wu}, \textbf{Caiming Xiong}, \textbf{Shafiq Joty} \\[0.5em]
    Salesforce AI Research \\
    {\texttt{\{austin.xu, cxiong, sjoty\}@salesforce.com}}\\[1em]
}}}

\iclrfinalcopy 
\begin{document}

\maketitle

\begin{abstract}
Finetuning specialized generative evaluators has emerged as a popular paradigm to meet the increasing demand for scalable evaluation during both training and test-time. However, recent work has largely focused on applying new methodology, such as reinforcement learning (RL), to training evaluators, shying away from large-scale, data-driven development. 
In this work, we focus on data scaling, curating a set of 2.5M samples spanning five unique evaluation tasks (pairwise, step-level, reference-free and reference-based verification, and single rating) and multiple domains focused on reasoning evaluation. With our data, we train Foundational Automatic Reasoning Evaluators (FARE), a family of 8B and 20B (with 3.6B active) parameter evaluators, with a simple iterative rejection-sampling supervised finetuning (SFT) approach. FARE-8B challenges larger specialized RL-trained evaluators and FARE-20B sets the new standard for open-source evaluators, surpassing specialized 70B+ evaluators. Beyond static benchmarks, we evaluate FARE in real-world tasks: As inference-time rerankers, FARE-20B achieves near-oracle performance on MATH. As verifiers in RL training, FARE improves the downstream RL-trained model performance by up to 14.1\% vs. string-matching verifiers. When initialized from FARE, a continually-finetuned FARE-Code outperforms gpt-oss-20B by 65\% on evaluating test-case quality.

{\centering
    \vspace{1em}
    \huggingface~~\href{https://huggingface.co/collections/Salesforce/fare-68eebc1779713d853886f67b}{\texttt{The FARE family of evaluators}}\\[0.5em]}

\begin{figure}[h!]
    \centering
    \includegraphics[width=0.8\linewidth]{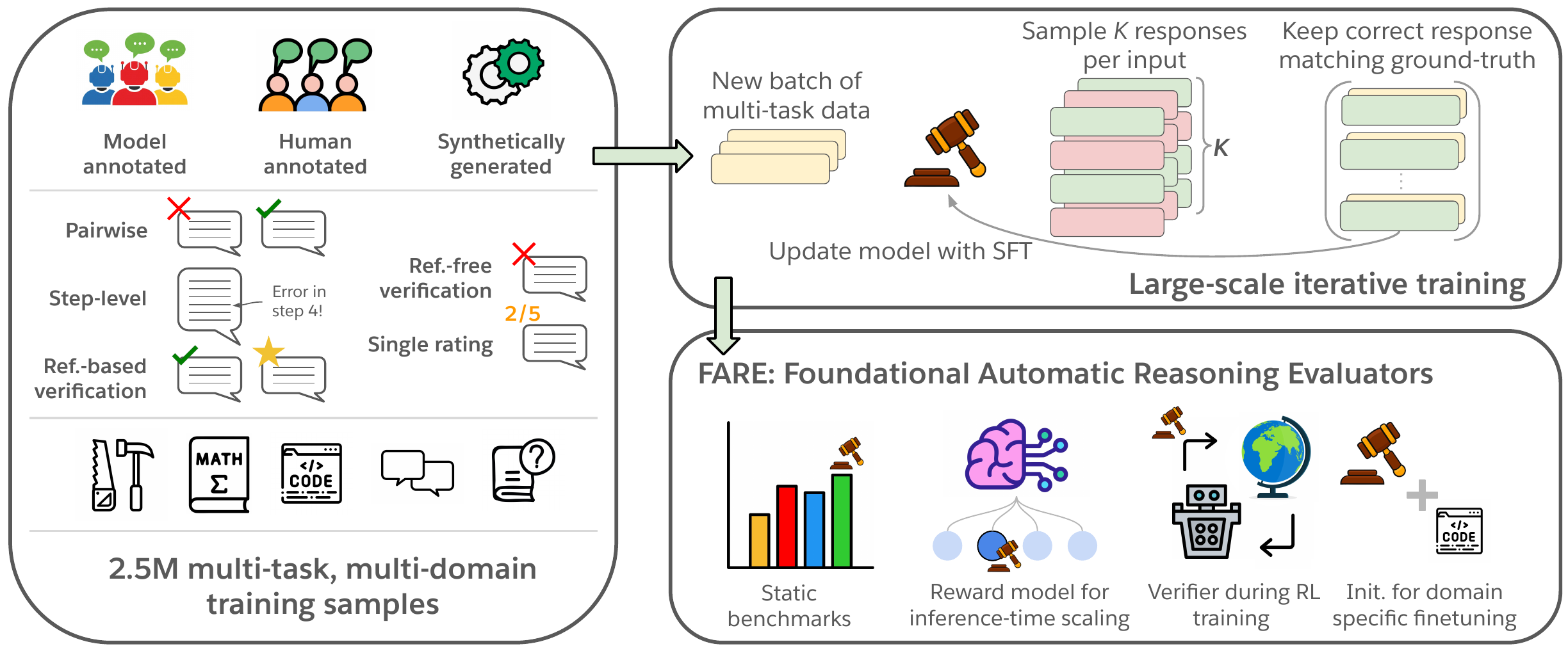}
    \caption{Overview of our work. We curate 2.5M multi-task, multi-domain training samples (left) and use large-scale iterative rejection sampling SFT to train \textbf{\ourmodel}, a family of automatic evaluators (top right). We evaluate \ourmodel on static benchmarks and on various real-world downstream tasks.}
    \label{fig:overview}
\end{figure}
\end{abstract}

\section{Introduction}
The past two years has seen the rapid adoption of large language models (LLMs) as automatic evaluators in response to demands for scalable evaluation of LLM outputs. LLM-based evaluators serve as judges for popular benchmarks~\citep{dubois2024length,zheng2023judging}, generative reward models for preference optimization~\citep{yuan2024self,wu2024meta}, and verifiers/critics in inference-time scaling settings~\citep{mcaleese2024llm,zhou2025evaluating}. The widespread integration of evaluators into nearly every phase of the LLM development cycle~\citep{wu2025sailing} demands evaluators that can handle multiple \textit{evaluation tasks} while operating effectively across diverse \textit{domains}.

Different settings require different evaluation abilities: Alignment needs evaluators capable of comparing different responses, i.e., pairwise evaluation, whereas monitoring model outputs requires finding minute mistakes, i.e., step-level evaluation. More recently, generative evaluators are tasked with providing reward signals during reinforcement learning (RL) training~\citep{team2025kimi,jiang2025pag}, and are expected to grow in importance as RL moves towards unverifiable domains~\citep{gunjal2025rubrics,jayalath2025compute} in complex reasoning settings~\citep{ke2025survey,ferrag2025llm}. As evaluators take central roles in training and evaluating the next generation of models, they must be flexible enough to evaluate as the setting demands. 

Compounding the challenges of multi-task evaluation is the expanding number of \textit{domains} requiring evaluation: RL-training has quickly moved from math reasoning (e.g.,~\citet{yu2025dapo}) to general-purpose reasoning~\citep{ma2025general} (e.g., history or economics). Agentic settings introduce additional wrinkles: With autonomously acting single agents~\citep{openai2025deepresearch,nguyen2025sfr,wei2025webagent} and complex multi-agent workflows~\citep{openmanus2025,alzubi2025opendeepsearch} now being set free to browse the web and act on behalf of users with minimal oversight, evaluators must assess not only agent reasoning, but also proposed tool-use. These systems, sometimes built with intricate \textit{model-generated}~\citep{hu2024automated,zhang2024aflow,ke2025mas} interdependencies, are bottlenecked, in part, by subpar evaluation~\citep{cemri2025multi}.

Unfortunately, recent work in the open-source automatic evaluation community has failed to meet these twin demands of \textit{multi-task, multi-domain} evaluators, opting instead in training task-specialized evaluators at relatively small data scales. We break this trend by \textit{scaling up data}, curating 2.5M multi-task, multi-domain training samples that emphasize reasoning settings. As shown in~\Cref{fig:overview}, our data mix covers 5 distinct tasks and various domains like math, code, tool-use evaluation, and natural language reasoning. With our data, we train, \textbf{Foundational Automatic Reasoning Evaluators (FARE)}, two best-in-class evaluators. As shown in~\Cref{fig:overview}, our contributions are 
\begin{itemize}[leftmargin=*,noitemsep]
    \item \textbf{Multi-task, multi-domain dataset:} We curate a large-scale, multi-task training set with an emphasis on reasoning-centric settings. We supplement existing human-and model-annotated data with synthetic data created from challenging new seed datasets.
    \item \textbf{Scalable learning via iterative rejection sampling:} We show that iterative rejection sampling supervised finetuning (RS-SFT) is a stable approach for training evaluators at scale. The semi-online nature of RS-SFT avoids problematic teacher model distribution shifts while bringing computationally stable and efficient model updates. Through ablations, we quantify the impact of training pipeline features like quantity of direct judgment data and the use of a continuous curriculum.
    \item \textbf{The \ourmodel family of evaluators:} We train \ourmodelsmall and \ourmodellarge and rigorously assess them with 7 challenging benchmarks and 3 practical downstream settings: test-time response reranking, RL-training verification, and domain-specific continual finetuning.
\end{itemize}
Our trained models are both well-rounded and high-performing. Out of the box, \ourmodel improve generator performance at test-time, achieving near oracle reranking performance on MATH, and provide clear rewards during general-domain RL training, boosting downstream performance by 14.1\% over typical string-matching verifiers. With minimal continual training, \ourmodel can be adapted to specific domains like code, beating gpt-oss-20B by 65\% in code test-case quality evaluation.
\section{Background and Related Work}\label{sec:background}
An automatic evaluator (AE) $\pi_\theta: \mathcal{X} \rightarrow \mathcal{Y}$ maps input $x = (p, q, \mathcal{R}) \in \mathcal{X}$ to output $y = (c, j) \in \mathcal{Y}$. Input $x$ consists of $p$, the \textit{evaluation protocol} that specifies both the \textit{evaluation task} (e.g., pairwise comparison, verification) and evaluation rubric, $q$, the original question, and $\mathcal{R}$, set of model responses to be evaluated. The output $y$ consists of a natural language critique $c$ and final judgment $j$. The AE may also be prompted to omit the critique $c$ and directly output the judgment $j$, which we denote as $y = (\emptyset, j)$. The specific \textit{evaluation protocol} $p$ determines the elements of set $\mathcal{R}$ and the exact form of judgment $j$. For example, in pairwise evaluation, $\mathcal{R}$ consists of two responses $\{r_1, r_2\}$ and the judgment is a choice between the two (``A'' or ``B''), whereas in single-rating, $\mathcal{R}$ consists of a single response $r$ and the judgment is an integer on a 1-5 scale. In this work, we focus on training automatic evaluators capable of the 5 evaluation tasks shown in~\Cref{fig:overview}:
\begin{itemize}[leftmargin=*,noitemsep,topsep=1pt]
    \item \textbf{Pairwise comparisons}: Given response set $\mathcal{R} = \{r_1, r_2\}$, the AE selects the better of $r_1$ and $r_2$. 
    \item \textbf{Step-level evaluation}: Given response set $\mathcal{R} = \{r_\text{[steps]}\}$, where $r_\text{[steps]}$ is a single model response broken down into steps, the AE identifies step-level errors.
    \item \textbf{Reference-based verification}: Given response set $\mathcal{R} = \{r_\text{cand}, r_\text{ref}\}$, where $r_{\text{cand}}$ is the candidate and $r_\text{ref}$ is the reference, the AE determines if $r_\text{cand}$ is correct based on $r_\text{ref}$.
    \item \textbf{Reference-free verification}: Given response set $\mathcal{R} = \{r\}$, the AE determines if $r$ is correct.
    \item \textbf{Single rating}: Given response set $\mathcal{R} = \{r\}$, the AE assigns an integer score to $r$.
\end{itemize}

\textbf{Past work in generative automatic evaluators.} Capable LLMs, like GPT-4, were originally prompted as scalable evaluators~\citep{wang2023chatgpt,liu2023g,fu2024gptscore,chiang2023can}. Subsequent analysis revealed pitfalls of prompted approaches, like biases with respect to position~\citep{wang2023large,li2023generative}, length~\citep{zeng2023evaluating,park2024offsetbias}, or self-preference~\citep{panickssery2024llm}. Finetuning specialized evaluators emerged as a result, with early approaches using teacher model outputs to do supervised finetuning (SFT)~\citep{kim2023prometheus,kim2024prometheus,li2023generative,park2024offsetbias,skyworkcritic2024} or direct preference optimization (DPO)~\citep{hu2024themis,ye2024beyond}, often focusing only on one or two evaluation tasks. More recent methods moved to reasoning models as teachers~\citep{khalifa2025process}.

\citet{vu2024foundational,wang2024direct,cao2024compassjudger,alexandru2025atla} train \textit{foundational evaluators} at larger data scales with multi-protocol capabilities via \textit{offline} training methods like SFT or DPO. Such approaches take inspiration from general-purpose, large-scale multi-task learning~\citep{sanh2021multitask,raffel2020exploring,wei2021finetuned}, which showed broad generalization capabilities emerge with the scaling of training data. Foundational evaluators likewise were empirically shown to generalize to unseen evaluation tasks, prompts, and criteria while being more robust to common biases~\citep{vu2024foundational,wang2024direct}.

Recent work has focused on \textit{methodological} advances, either using inference-time scaling~\citep{liu2025inference,chan2025j1,zhao2025genprm} or \textit{online} training like reinforcement learning from verifiable rewards (RLVR)~\citep{chen2025judgelrm,chen2025rmr1rewardmodelingreasoning,whitehouse2025j1,xu2025j4r,xiong2025stepwiser} to improve evaluator performance. Because RLVR is computationally demanding with relatively brittle training pipelines \citep{dsr1_guo2025deepseek,yang2025qwen3}, recent evaluators are typically trained on a small amount of data for a single task.
Our work bridges early work in training foundational evaluators with more recent methodological advancements, demonstrating that a simple semi-online training approach enables stable multi-task training at scale.

\textbf{Desiderata for a new generation of evaluators.} Here, we outline our design philosophy for \ourmodel. Beyond accuracy and robustness, we seek \textit{efficiency}, as many evaluation settings like inference-time reranking or RL rollout verification demand low latency. In contrast to recent long chain-of-thought (CoT) evaluators~\citep{chen2025rmr1rewardmodelingreasoning,khalifa2025process}, we select base models with either no or very compact ``thinking'' CoTs. We also explicitly \textit{avoid having the evaluator generate reference answers}. Past work has used evaluators to generate references during evaluation~\citep{zheng2023judging,li2024crowdsourced} or training rollout~\citep{chen2025rmr1rewardmodelingreasoning}. Not only does this risk severely degrading performance when the reference is wrong~\citep{krumdick2025no}, it also converts the relatively easier evaluation task into a harder generation task~\citep{zhou2025variation}.

\section{FARE: data and training recipe}\label{sec:method}

\subsection{Data curation}\label{sec:method:data}
We use two data approaches for curating our final training mix: Using \textbf{\texttt{Existing}} high quality training datasets created for evaluator and preference finetuning
and generating \textbf{\texttt{Synthetic}} datasets through programmatic error injection and a generate-then-grade strategy.

\textbf{\texttt{Existing}} data consists of training samples from proven sources that have produced effective evaluators~\citep{vu2024foundational,cao2024compassjudger}. These datasets consist of high quality annotations from humans and frontier LLMs, and cover evaluation tasks like step-level, single-rating, and pairwise evaluation and domains like chat quality, code, and safety. Following~\cite{wang2024direct}, we largely focus our data collection on \textit{modern} (2024 and beyond) datasets, as these datasets contain the most up-to-date model responses and fresh annotations. Beyond evaluator-specific training, we take advantage of existing preference fine-tuning datasets used for RLHF \citep{instructgpt_ouyang2022training} and DPO training \citep{dpo_rafailov2023direct}, converting these directly into pairwise evaluation samples. In domains with objectively correct answers, e.g., math, we also create verification training data with positive responses as correct/reference responses and negative responses as incorrect responses. 

We hand-craft evaluation rubrics for each source dataset that follow annotation instructions given to human annotators or models, if existing. If such original instructions do not exist, we write custom evaluation rubrics for each source dataset based on the data composition and domain.~\Cref{app:rubric_example} provides an example rubric. \textbf{\texttt{Existing}} data lays a solid foundation, with 1.4M samples already dwarfing data scales found in recent work. However, upon inspection, we found three clear shortcomings: (1) Newly relevant tasks, like verification, were underrepresented. (2) Existing pairwise task data focused largely on chat-related topics and less-so on reasoning-relevant domains. (3) Questions and responses from newer, challenging datasets produced to meet the needs of reasoning-focused RL training were absent. To address these limitations, we supplement with synthetic data.

\begin{figure}
    \centering
    \includegraphics[width=0.9\linewidth]{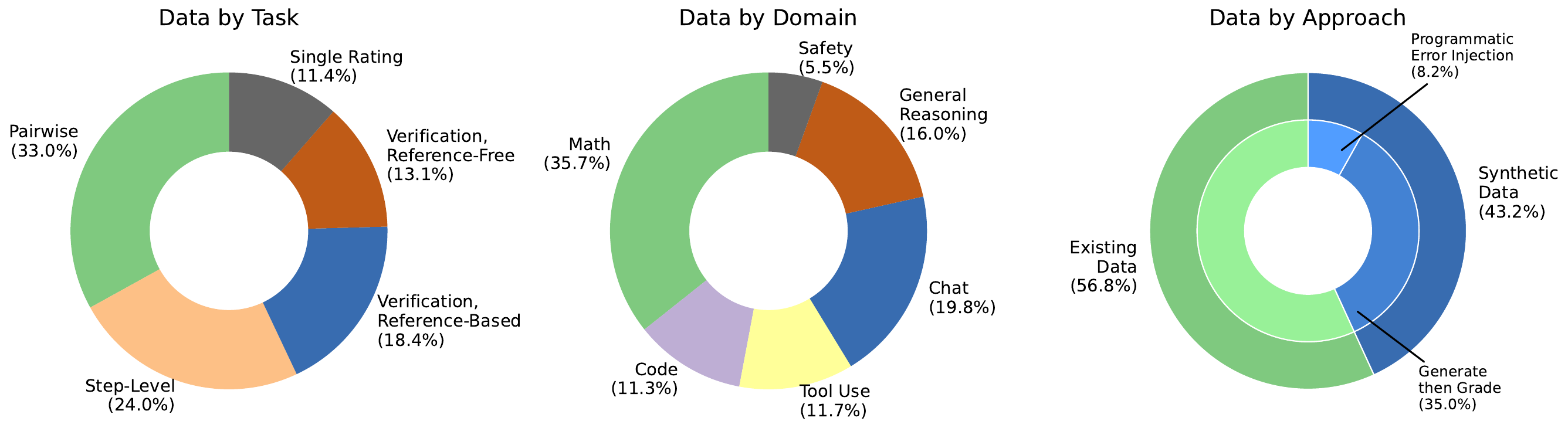}
    \caption{Breakdown of our curated training dataset of 2.5M samples by task (left), domain (center), and curation approach (right). Domain breakdown excludes step-level data, which is entirely math.}
    \label{fig:dataset}
\end{figure}

\textbf{\texttt{Synthetic}} data is generated from a diverse set of challenging seed datasets using two approaches:
\begin{itemize}[leftmargin=*,noitemsep,topsep=5pt]
    \item \textbf{Programmatic error injection.} We employ programmatic error injection when applicable, such as tool-use and function-calling data. For example, to create pairwise tool-use data, we inject errors (e.g., type error, extra argument, syntax errors) in correct function calls. \austin{New, tying back to end of last paragraph:}This approach increases the amount of tool-use evaluation data, adding both pairwise and verification samples.
    \item \textbf{Generate-then-grade.} Here, we leverage a mix of recent and established training datasets comprised of question $q$ and verifiable ground-truth answer $a$. We sample up to $20$ responses per $q$ from various generator models, then grade each responses based on $a$. After grouping responses by correctness, we create verification and pairwise samples. We use 12 unique generators from 6 model families, covering reasoning and non-reasoning models. This ensures that \ourmodel are trained on a diverse array of model responses, enabling better generalization across distinct response distributions~\citep{singh2025shelf}. \austin{New, tying back to end of last paragraph:}Generate-then-grade enables us to incorporate problems from recent, challenging datasets covering frontier math and reasoning tasks. This enhances the quality of our pairwise and verification data with difficult-to-evaluate reasoning-focused samples.
\end{itemize}
Our final dataset comprises 2.5M training samples; an overview is shown in~\Cref{fig:dataset}, with breakdowns by task, domain, and curation approach, and exact dataset sources are described in~\Cref{app:data}.

\subsection{Model training}\label{sec:method:training}
\textbf{General training recipe.} We aim to train an automatic evaluator $\pi_\theta$, which we call the \textit{policy model} parametrized by $\theta$. Our data curation process yields a training dataset of $N$ samples $\mathcal{D} = \{(x_i, j_i^\star)\}$, where $j_i^\star$ denotes the ground-truth judgment. Corresponding ground truth critiques $c^\star$ are not typically available in evaluator training data. As a result, past work has resorted to offline teacher-model based approaches or online RL-based approaches, as discussed in~\Cref{sec:background}.

These established approaches have their limitations: Teacher models introduce distribution shifts with respect to the policy model, a common problem with imitation learning \citep{ross2011reduction}, with choice of teacher model having a large impact on downstream performance~\citep{guha2025openthoughts}. On the other hand, RL training is compute and time-intensive, making it difficult to scale to large data quantities, with past work~\citep{liu2025prorl} only exploring small (1.5B parameter) models via ad-hoc interventions like reference policy resetting. 

We borrow desirable qualities from both paradigms and use semi-online iterative rejection sampling SFT (RS-SFT)~\citep{touvron2023llama,dong2023raft}, which was recently shown to be competitive with RLVR approaches~\citep{xiong2025minimalist}. RS-SFT avoids sub-optimal distribution shift by finetuning on correct evaluation traces produced by the policy model while employing a computationally lightweight policy update step. This enables simple yet stable scaling to millions of training samples \textit{without} sampling from a teacher model. An overview of our approach is shown in~\Cref{fig:overview}.

Concretely, we train $\pi_\theta$ as follows. We split our $N$ training samples into disjoint rollout batches $\mathcal{B}_t = \{(x_{i,t}, j_{i,t}^\star)\}$ of fixed size $N_\text{rollout}$ and  initialize the initial policy $\pi_{\theta_0}$ to be an existing post-trained LLM (e.g., gpt-oss-20B). Then for step $t = 0, \ldots, T-1$, we perform the following:
\begin{itemize}[leftmargin=*,noitemsep,topsep=5pt]
    \item \textbf{Rollout from previous policy}: For inputs $x_{i,t}$ from $\mathcal{B}_t$, sample $K$ responses per input from policy $\pi_{\theta_t}$, denoted $\{\hat{y}_{i,t}^{(1)}, \ldots, \hat{y}_{i,t}^{(K)}\}$.
    \item \textbf{Rejection sampling}: For each of the $K$ responses $\{\hat{y}_{i,t}^{(1)}, \ldots, \hat{y}_{i,t}^{(K)}\}$, we determine correctness with ground-truth judgment $j_{i,t}^\star$. For inputs with correct responses, one randomly chosen response is kept. Any inputs without correct responses are discarded. We denote the collected set of inputs and corresponding correct responses as $\mathcal{D}_t$.
    \item \textbf{Policy update}: Use $\mathcal{D}_t$ to update the policy weights via SFT, initializing with $\theta_t$:
    \begin{align}
        \theta_{t+1} = \argmax_{\theta} \sum\limits_{(x, y) \in \mathcal{D}_t} \log\pi_{\theta}(y | x)
    \end{align}
\end{itemize}
Our approach draws inspiration from algorithms such as STaR~\citep{zelikman2022star} and RAFT~\citep{dong2023raft}, with some key differences. STaR notably re-initializes training from $\pi_{\theta_0}$ for each iteration $t$ and samples only one greedy response per input prompt, while RAFT relies on an external reward model to rank generated outputs. Because the automatic evaluation setting is inherently verifiable, we omit the need for a reward model to rank sampled responses. 

Specific to evaluators, the Self-Taught Evaluator (STE) paradigm of~\citet{wang2024self} and follow-up EvalPlanner~\citep{saha2025learning} are closely related to our approach. STE follows STaR with policy re-initialization and EvalPlanner uses multiple SFT and DPO training runs per iteration $t$. Further, these works use in-the-loop synthetic data generation, sampling responses to a small number ($<25$K) of seed questions from a \textit{fixed} generator. Pairwise samples are then created from correct/incorrect responses and used to train the model. This data generation approach cannot be adapted to create data for other tasks, like step-level evaluation, fundamentally limiting the task abilities of STE. A secondary concern is a lack of exposure to diverse response distributions; Evaluations in~\citet{wang2024direct} show that scaling training data with a simpler training recipe leads to better generalization across benchmarks compared to STE. 

\textbf{Batch composition.} For each rollout batch $\mathcal{B}_t$, we select unseen training samples from our curated training dataset, ensuring the task mixture is consistent with global task composition. For example, 33\% of our overall training data are pairwise tasks (\Cref{fig:dataset}), so 33\% of the input prompts in $\mathcal{B}_t$ are sampled from unseen pairwise samples. 
We then sample $K=4$ responses per sample with a temperature of $0.9$, and determine correctness based on final judgment. 

\textbf{Inclusion of direct judgment data.} Past work~\citep{wang2024direct,cao2024compassjudger} has showed the importance of including \textit{direct judgment} data samples to isolate judgment training signal. These are samples where the critique $c$ is omitted, and the input protocol $p$ is modified to prompt directly for a judgment. To precisely control the fraction of direct judgment data, we convert a fixed percent of $\mathcal{D}_t$ to direct judgment data\austin{on a per-task basis} by dropping generated critiques and modifying the input prompt accordingly. In~\Cref{sec:exp:analysis}, we ablate the proportion of direct judgment data and show such data enables \ourmodel to be prompted to exclude critiques for faster inference. 

\textbf{Per-batch continuous curriculum learning.} We additionally use a \textit{continuous curriculum} in training: For each $(x,y) \in \mathcal{D}_t$, we compute the pass percentage from the $K=4$ rollout generations for $x$, then sort the dataset in descending order of pass percentage. That is, samples where all $4$ sampled outputs are correct are used to update the model first, and samples where only $1$ of $4$ sampled outputs are correct are used to update the model last. We find this has negligible impact on pairwise domains but large impacts in step-level evaluation, as we show in~\Cref{sec:exp:analysis}. 

\textbf{Base models.} We train two models starting from Qwen3-8B-Base~\citep{yang2025qwen3} and gpt-oss-20B~\citep{agarwal2025gpt}, denoted \ourmodelsmall and \ourmodellarge. We find Qwen3-8B (post-trained) to be over-trained, and therefore cold-start Qwen3-8B-Base from SFT data from Qwen2.5-32B-Instruct, which we denote Qwen3-8B-ColdStart. See~\Cref{app:data:training} for additional details.
\begin{table}[t!]
\caption{Pairwise evaluation results, with \textbf{best} and \underline{second-best} performance in each section marked. \ourmodel achieve best-in-class performance, even outperforming frontier models in tool-use evaluation. $\dagger$ indicates that benchmark uses consistent accuracy (25\% random baseline). 
}
\label{tab:pairwise}
\centering
\groupedRowColors{5}{16}{lightgray!50}{lightgray!50}
\resizebox{0.7\textwidth}{!}{%
\begin{tabular}{ l !{\vrule width -1pt}c !{\vrule width -1pt}c !{\vrule width -1pt}c !{\vrule width -1pt}c !{\vrule width -1pt}c !{\vrule width -1pt}l }
\toprule
& JudgeBench$^\dagger$ & RJB$^\dagger$ & PPE Correctness & RM-Bench & When2Call$^\dagger$ \\ 
\midrule
RISE-Judge-7B & 44.57 &  34.73  & 61.3  &   77.2 &  47.22\\
EvalPlanner-8B    & 30.20 &  - &  52.8  &   68.1 &  -    \\
J1-8B   &   42.00 &  - &  59.2  &   73.4 &  -    \\
RM-R1-14B    &  46.86 &  \underline{43.70}  & \textbf{64.0}  &   \underline{79.6} &  19.89\\
CompassJudger-7B  & 49.14 &  37.76  & 60.9  &   \textbf{82.2} &  41.67\\
Atla Selene 8B    & 21.14 &  12.41  & 53.3  &   71.9 &  \underline{56.00}\\
CompassJudger-14B & \underline{50.29} &  37.69  & 62.0  &   77.7 &  44.56\\
\textbf{\ourmodelsmall}  & \textbf{55.71} &  \textbf{51.05}  & \underline{63.8}  &   79.2 &  \textbf{80.33}\\ 
\midrule
RISE-Judge-32B    & 46.86 &  42.35  & 63.5  &   82.2 &  46.44\\
CompassJudger-32B & 54.57 &  \underline{46.53}  & 65.6  &   80.1 &  \underline{51.89}\\
RM-R1-32B    &  54.29 &  46.39  & 65.9  &   81.5 &  23.89\\
EvalPlanner-70B   & 56.60 &  - &  70.2  &   82.1 &  -    \\
J1-70B  &   60.00 &  - &  \underline{72.8}  &   \underline{82.7} &  -    \\
\textbf{\ourmodellarge} & \textbf{64.29} &  \textbf{57.05}  & \textbf{74.4}  &   \textbf{90.5}    & \textbf{76.67}\\ 
\midrule
{Qwen3-8B-ColdStart} & 48.29 & 40.59 & 60.5 & 78.07 & 59.67 \\
Qwen3-8B & 52.27 & 43.56 & 64.8 & 79.9 & 64.78 \\
gpt-oss-20B & 59.43 & 50.51 & 71.7 & 89.9 & 61.33 \\
gpt-oss-120B & 70.29  & 58.26 & 77.8 & 92.0 & 70.00 \\
GPT-5-nano & 59.71 & 51.52 & 80.7 & 92.3 & 50.02\\
GPT-5 & 84.86  & 79.57 & 87.0 & 93.8 & 75.78 \\
\bottomrule
\end{tabular}%
}
\end{table}

\begin{table}[t!]
\caption{ProcessBench results, with \textbf{best} and \underline{second-best} performance in each section marked. \ourmodellarge almost matches GPT-5 with the same prompt, achieving best-in-class performance, while \ourmodelsmall beats comparably sized generative (Gen.) evaluators.}
\label{tab:process}
\centering
\groupedRowColors{5}{20}{lightgray!50}{lightgray!50}
\resizebox{0.7\textwidth}{!}{%
\begin{tabular}{llccccc}
\toprule
\multicolumn{2}{l}{} & GSM8K & MATH & OlympiadBench & OmniMATH & Overall \\ \midrule
PRM & SkyworkPRM-1.5B & 59.0 & 48.0 & 19.3 & 19.2 & 36.4 \\
PRM & Math Shepherd-7B & 47.9 & 29.5 & 24.8 & 23.8 & 31.5 \\
PRM & SkyworkPRM-7B & 70.8 & 53.6 & 22.9 & 21.0 & 42.1 \\
PRM & ActPRM-7B & \underline{82.7} & \textbf{82.0} & \underline{72.0} & \underline{67.3} & \underline{76.0} \\
PRM & Qwen2.5-7B-PRM800K & 68.2 & 62.6 & 50.7 & 44.3 & 56.5 \\
PRM & Qwen2.5-Math-7B-PRM & 82.4 & 77.6 & 67.5 & 66.3 & 73.5 \\
PRM & Qwen2.5-Math-72B-PRM & \textbf{87.3} & \underline{80.6} & \textbf{74.3} & \textbf{71.1} & \textbf{78.3} \\
\midrule
Gen. & Qwen2.5-Math-7B & 26.8  & 25.7 & 14.2 & 12.7 & 19.9 \\
Gen. & RL Tango-7B & 53.1 & 48.2 & 37.8 & 36.3 & 43.9 \\
Gen. & StepWiser-1.5B  & 46.9 & 43.4 & 26.3 & 28.4 & 36.3 \\
Gen. & StepWiser-7B  & \textbf{72.4} & \textbf{68.3} & \underline{54.4} & \underline{52.4} & \underline{61.9} \\
Gen. & \textbf{\ourmodelsmall} & \underline{68.5} & \underline{67.7} & \textbf{59.9} & \textbf{58.1} & \textbf{63.5} \\ 
\midrule
Gen. & Llama-3.3-70B & 82.9 & 59.4 & 46.7 & 43.0 & 58.0 \\
Gen. & Qwen2.5-Coder-32B & 68.9 & 60.1 & 48.9 & 46.3 & 56.1 \\
Gen. & QwQ-32B & \underline{88.0} & \underline{78.7} & \underline{57.8} & \underline{61.3} & \underline{71.5} \\
Gen. & Qwen2.5-Math-72B & 65.8 & 52.1 & 32.5 & 31.7 & 45.5 \\
Gen. & GPT-4o & 79.2 & 63.6 & 51.4 & 53.5 & 61.9 \\
Gen. & \textbf{\ourmodellarge} & \textbf{89.8} & \textbf{87.8} & \textbf{80.0} & \textbf{79.9} & \textbf{84.4} \\ 
\midrule
Gen. & Qwen3-8B-ColdStart & 37.0 & 41.0 & 36.3 & 38.9 & 38.3 \\ 
Gen. & Qwen3-8B & 63.2 & 64.0 & 51.5 & 48.2 & 56.7 \\ 
Gen. & gpt-oss-20B & 79.3 & 79.4 & 68.8 & 68.2 & 73.9 \\ 
Gen. & gpt-oss-120B & 89.6 & 87.6 & 80.8 & 76.0 & 83.5 \\ 
Gen. & GPT-5-nano & 83.8 & 87.0 & 80.6 & 77.1 & 82.1 \\ 
Gen. & GPT-5 & 91.4 & 89.5 & 80.6 & 76.9 & 84.6 \\ 
\bottomrule
\end{tabular}
}
\vspace{-1\baselineskip}
\end{table}

\section{Experiments}
We evaluate \ourmodel on both \textit{core benchmarks}, static benchmarks for automatic evaluators, and in \textit{downstream settings}, which simulate real applications of evaluators. We provide descriptions of benchmarks and baselines in~\Cref{app:benchmarks} and additional ablations and analysis in~\Cref{sec:exp:analysis}.

\subsection{Core benchmarks}\label{sec:exp:core}
\textbf{Setup.} We evaluate along five diverse aspects: \Ni \textit{reasoning} with JudgeBench~\citep{tan2024judgebench}, ReasoningJudgeBench (RJB)~\citep{xu2025j4r}, and PPE Correctness~\citep{frick2024evaluate}, \Nii \textit{bias and robustness} with RM-Bench~\citep{liu2024rm}, \Niii \textit{tool-use} with When2Call~\citep{ross2025when2call}, \Niv \textit{step-level error identification} with ProcessBench~\citep{zheng2024processbench}, and \Nv  \textit{reference-based verification} with VerifyBench~\citep{yan2025verifybench}. 

For RM-Bench and PPE Correctness pairwise benchmarks, we adopt default evaluation setups, running each benchmark once with a fixed random ordering of responses. For other pairwise benchmarks, we report \textit{consistent-accuracy}~\citep{tan2024judgebench}, where each test sample is run twice, swapping the order of response A and response B. If the evaluator selects a different response between runs (i.e., is positionally biased), then the sample is marked incorrect; if the evaluator is consistent, the judgment is graded against the ground-truth. For ProcessBench and VerifyBench, we report F1-score\footnote{
    ProcessBench defines their reported F1-score differently from the traditional F1-score; See \href{https://github.com/QwenLM/ProcessBench/issues/1}{this link}.
} and accuracy, respectively. We compare \ourmodel against other finetuned generative and prompted evaluators. We report official numbers from past benchmarks, reporting sources in~\Cref{app:benchmarks}. If necessary, we run each baseline using its own prompt template. For ProcessBench, we additionally compare against non-generative process reward models (PRMs).

\textbf{Results.} \Cref{tab:pairwise,tab:process,tab:verification} present our results on pairwise, step-level, and reference-based verification benchmarks, respectively. Our prompts are provided in~\Cref{app:prompts_examples}.

\begin{wraptable}{r}{7.5cm}
\caption{Ref.-based verification, with \textbf{best} and \underline{second-best} performance marked per section. \ourmodel beat general-purpose verifiers in hard settings.
}
\label{tab:verification}
\centering
\groupedRowColors{6}{15}{lightgray!50}{lightgray!50}
\resizebox{0.45\textwidth}{!}{%
\begin{tabular}{lcc}
\toprule
 & VerifyBench & VerifyBench-Hard \\ 
\midrule
Math-Verify & 45.90 & 32.50 \\
\midrule
GPT-4o mini & \underline{92.85} & 72.30 \\
Llama-3.1-8B & 73.05 & 43.20 \\
Qwen3-4B & 92.00 & \underline{72.40} \\
Phi-4 & 89.35 & 56.60 \\
Yi-1.5-9B & 87.70 & 61.40 \\
\textbf{\ourmodelsmall} & \textbf{93.20} & \textbf{78.40} \\ 
\midrule
GPT-4o & 93.15 & \underline{72.60} \\
Llama-4-Scout & 90.01 & 48.50 \\
Llama3.3-70B & 83.25 & 54.70 \\
Qwen2.5-72B & 92.35 & 62.40 \\
Qwen3-32B & \textbf{95.80} & 71.80 \\
\textbf{\ourmodellarge} & \underline{94.95} & \textbf{85.10} \\ 
\midrule
Qwen3-8B-ColdStart & 92.45 & 72.60 \\
Qwen3-8B & 94.00 & 70.90 \\
gpt-oss-20B & 91.95 & 83.60 \\
gpt-oss-120B & 95.35 & 88.30 \\
GPT-5-nano & 94.65 & 84.00 \\
GPT-5 & 96.10 & 90.50 \\
\bottomrule
\end{tabular}%
}
\vspace{-\baselineskip}
\end{wraptable}

\Cref{tab:pairwise} shows that across diverse pairwise benchmarks, \ourmodel exhibit best-in-class performance, outperforming comparably sized baselines. \ourmodelsmall is the strongest small judge, outperforming recently released RL-trained models like J1-8B and RM-R1-14B by 13.71 and 6.57 absolute points on JudgeBench, respectively. \ourmodellarge challenges strong judges at 20B parameters, outperforming dense 70B-sized judge models despite having 3.5x fewer total parameters and nearly 20x fewer active parameters. The strong performance of \ourmodel across reasoning benchmarks, which span math to scientific domains to causal reasoning, show that our models excel at discerning between objectively correct and incorrect responses.
Beyond reasoning settings, \ourmodel are generally robust to subtle, stylistic biases (RM-Bench) while excelling in tool calling evaluation (When2Call), which is increasingly important as agentic workflows grow in popularity.

\ourmodel also are extremely strong step-level evaluators, as shown by ProcessBench performance in~\Cref{tab:process}. \ourmodelsmall is the best small-sized generative critic model, outperforming the recently released StepWiser-7B~\citep{xiong2025stepwiser}, a RL-trained specialized step-level evaluator, by 1.6 points. \ourmodellarge outperforms the specialized Qwen2.5-Math-72B-PRM by 6.1 points, even matching GPT-5. Most notably, \ourmodel excel on the two most challenging splits, OlympiadBench and OmniMATH, with \ourmodelsmall beating StepWiser-7B by 5.6 points and \ourmodellarge beating Qwen2.5-Math-72B-PRM by 7.3 points on average.
Overall, \ourmodel are not only capable outcome-level evaluators, but are able to find subtle mistakes that manifest at the step-level.

\ourmodel are also capable verifiers, as shown in~\Cref{tab:verification}, with both models outperforming all reported baselines on VerifyBench-Hard~\citep{yan2025verifybench}. In particular, \ourmodellarge excels beats the next best model, GPT-4o, by 12.5 absolute points. As we demonstrate in~\Cref{sec:exp:applied}, when used as verifiers during GRPO settings, \ourmodel bring tangible benefits over typical string-matching verifiers.

\textbf{Scaling inference-time compute.} Recently, sampling parallel judgments and aggregating via majority vote, i.e., self-consistency~\citep{wang2022self}, has been used to improve evaluator performance. In~\Cref{tab:inference-time-scaling}, we use self-consistency with 32 responses (SC@32) on PPE, comparing with J1 and DeepSeek-GRM~\citep{liu2025inference}. DeepSeek-GRM also trains a MetaRM to perform judgment re-ranking at test-time. Across most splits, using SC@32 improves performance, with up to a 3.3 point improvement for \ourmodellarge on the MMLU-Pro split. Even without SC@32, \ourmodelsmall beats DeepSeek-GRM-27B + MetaRM, with the gap widening with extra compute. Similarly, the gap between \ourmodellarge and J1-70B grows with extra compute. Interestingly, we see that across both our models, performance on MBPP+ slightly degrades, indicating that SC may not be the optimal way to use compute across all domains. Nonetheless, we observe gains in the aggregate.

\begin{table}[t!]
\caption{Scaling inference-time compute for \ourmodel typically brings additional gains in performance: The performance gaps between \ourmodelsmall/DeepSeek-GRM and \ourmodellarge/J1-70B widens.
\shafiq{We should report the base (no TTS) perf.} \austin{This adds 5 rows to the table, we could report detailed results in appendix and ref there?}
\shafiq{we can possibly put them in the same row with an $\rightarrow$ sign?}\austin{good idea, done} }
\label{tab:inference-time-scaling}
\centering
\resizebox{0.95\textwidth}{!}{%
\begin{tabular}{lcccccc}
\toprule
PPE & MMLU-Pro & MATH & GPQA & MBPP+ & IFEval & Overall \\ 
 \midrule
J1-8B w/ SC@32  & \textcolor{gray}{65.6 $\rightarrow$} 67.5 & \textcolor{gray}{70.0 $\rightarrow$}  76.6 & \textcolor{gray}{53.2 $\rightarrow$} 55.7 & \textcolor{gray}{53.1 $\rightarrow$} 54.6 & \textcolor{gray}{ 54.0 $\rightarrow$} 54.9 & \textcolor{gray}{59.2 $\rightarrow$} 61.9 \\
J1-70B w/ SC@32 &  \textcolor{gray}{79.0 $\rightarrow$} 79.9 & \textcolor{gray}{86.0 $\rightarrow$} 88.1 & \textcolor{gray}{65.9 $\rightarrow$} 66.5 & \textcolor{gray}{\underline{66.0} $\rightarrow$} \textbf{66.5} & \textcolor{gray}{67.3 $\rightarrow$} 67.2 & \textcolor{gray}{72.8 $\rightarrow$} 73.6 \\
DeepSeek-GRM-27B w/ SC@32 & \textcolor{gray}{64.8$\rightarrow$} 65.5 & \textcolor{gray}{68.8 $\rightarrow$} 69.4 & \textcolor{gray}{55.6 $\rightarrow$} 56.0 & \textcolor{gray}{50.1 $\rightarrow$} 49.9 & \textcolor{gray}{59.8 $\rightarrow$} 61.0 & \textcolor{gray}{59.8 $\rightarrow$} 60.4 \\
DeepSeek-GRM-27B w/ MetaRM@32 & \textcolor{gray}{64.8$\rightarrow$} 68.1 & \textcolor{gray}{68.8 $\rightarrow$} 70.0 & \textcolor{gray}{55.6 $\rightarrow$} 56.9 & \textcolor{gray}{50.1 $\rightarrow$} 50.8 & \textcolor{gray}{59.8 $\rightarrow$} 70.4 & \textcolor{gray}{59.8 $\rightarrow$} 63.2 \\ 
\midrule
\ourmodelsmall w/ SC@32 & \textcolor{gray}{69.3 $\rightarrow$} 70.8 & \textcolor{gray}{79.7 $\rightarrow$} 80.4 & \textcolor{gray}{58.4 $\rightarrow$} 58.4 & \textcolor{gray}{55.7 $\rightarrow$} 54.9 & \textcolor{gray}{55.9 $\rightarrow$}  56.7 & \textcolor{gray}{63.8 $\rightarrow$} 64.2 \\
\ourmodellarge w/ SC@32 & \textcolor{gray}{\underline{80.3} $\rightarrow$} \textbf{83.6} & \textcolor{gray}{\underline{94.6} $\rightarrow$} \textbf{97.3} & \textcolor{gray}{\underline{68.5} $\rightarrow$} \textbf{71.1} & \textcolor{gray}{59.3 $\rightarrow$} 57.3 & \textcolor{gray}{\underline{69.1} $\rightarrow$}  \textbf{71.1} & \textcolor{gray}{\underline{74.4} $\rightarrow$} \textbf{76.6} \\
\bottomrule
\end{tabular}%
\vspace{-2\baselineskip}
}
\end{table}

\subsection{Downstream evaluation}\label{sec:exp:applied}
\textbf{Setup.} We apply \ourmodel on 3 downstream tasks: \Ni Reward model for inference-time scaling, \Nii verifier for GRPO training, and \Niii initialization for continual finetuning for domain-specific evaluation. We provide detailed explanations of our downstream evaluation settings in~\Cref{app:benchmarks:applied}.

\begin{figure}[b!]
    \vspace{-1\baselineskip}
    \centering
    \includegraphics[width=0.85\linewidth]{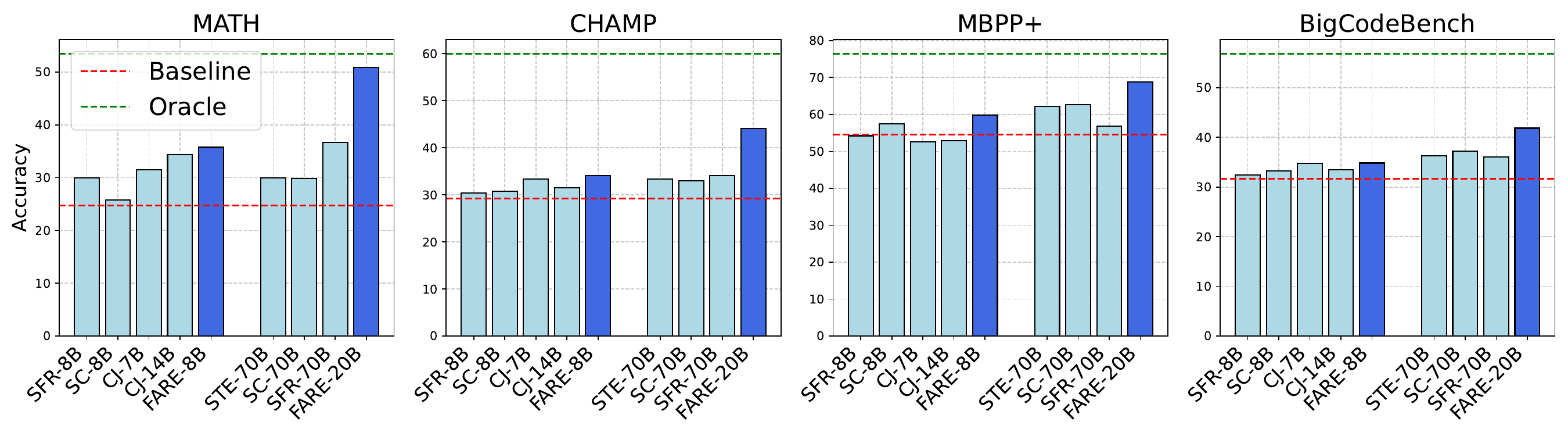}
    \caption{Best-of-10 performance for Llama-3.1-8B generator across 4 challenging benchmarks with baseline (red line) and oracle (green line) performance with \ourmodel and SFR-Judge (SFR), Skywork Critic (SC), Compass-Judger (CJ), and Self-Taught Evaluator (STE) as baselines. \ourmodel are the best small ($\leq$14B) and large ($\geq$ 20B) reranking models: \ourmodellarge achieves near oracle re-ranking performance on MATH, while \ourmodelsmall matches 70B judges.}
    \label{fig:jetts}
\end{figure}

\textbf{Reward model for inference-time scaling.} We use the standardized setup in JETTS~\citep{zhou2025evaluating}, which provides a set of 10 outputs from various generators and various benchmarks with corresponding correctness labels. Here, automatic evaluators rerank the responses, and performance is measured as the final performance of the evaluator-selected responses. We select the four most challenging benchmarks used in JETTS: MATH~\citep{hendrycks2021measuring}, CHAMP~\citep{mao2024champ}, MBPP+~\citep{evalplus}, and BigCodeBench~\citep{zhuo2024bigcodebench}. We compare against strong judge models benchmarked previously on JETTS: SFR-Judge-8B,70B~\citep{wang2024direct}, Skywork-Critic-8B,70B~\citep{skyworkcritic2024}, Self-Taught-Evaluator-70B~\citep{wang2024self}, and CompassJudger-7B,14B~\citep{cao2024compassjudger}. We utilize the default pairwise reranking setup and prompt \ourmodel to produce a judgment directly without explanation, i.e., $y = (\emptyset, j)$.

\Cref{fig:jetts} shows best-of-10 performance with Llama-3.1-8B as generator with baseline greedy (red line) and oracle re-ranking performance (green line). \ourmodel produce best-in-class reranking performance, with \ourmodelsmall roughly matching the performance of larger (70B) judges in math settings and outperforming all similar sized judges in coding domains. \ourmodellarge excels in math domains, \textit{approaching oracle-level reranking performance on MATH}, beating SFR-Judge-70B and Skywork-Critic-70B by 14 and 21 absolute points, respectively. Similarly, \ourmodellarge beats 70B+ judges by large margins in challenging coding domains. As we show in~\Cref{app:additional_results:jetts}, \ourmodel improve the performance of other generators and \ourmodellarge improves significantly over gpt-oss-20B as a reranker.

\textbf{Verifer for GRPO training.} We train with WebInstruct-Verified~\citep{ma2025general}, a multi-domain reasoning dataset, covering math, chemistry, etc. Verifier impact is measured via the downstream performance of the trained policy model on a fixed evaluation suite of MMLU-Pro \citep{wang2024mmlu}, GPQA-Diamond~\citep{rein2024gpqa}, MATH-500, Minerva-Math~\citep{lewkowycz2022solving}, OlympiadBench~\citep{he2024olympiadbench}, and AIME24 (Avg@32). We start from Qwen2.5-7B-Base~\citep{yang2024qwen2} with the default reward setup as~\citet{ma2025general} (see~\Cref{app:benchmarks:applied}), and we compare against training with the string-matching and trained verifier (General-Verifier) from~\citet{ma2025general}.

\begin{figure}
    \centering
    \includegraphics[width=0.7\linewidth]{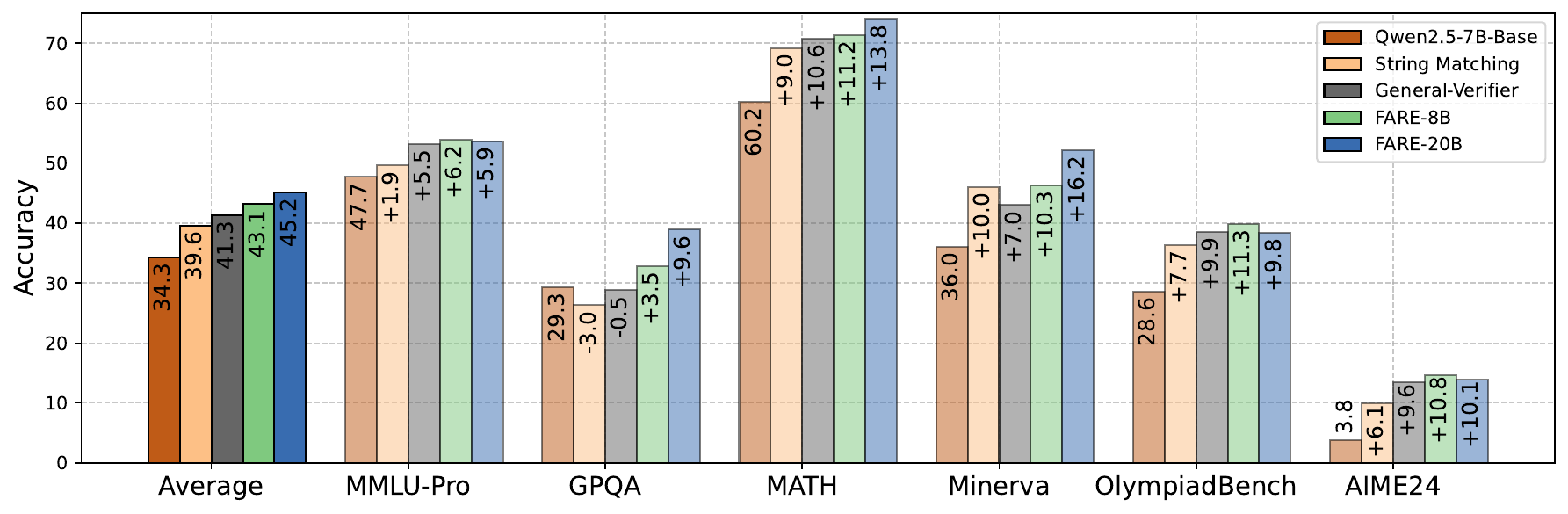}
    \caption{Performance of downstream GRPO-trained model with \ourmodel as reference-based verifiers. Moving from string based output matching or weaker verifiers to FARE brings tangible performance gains across both natural-language based (e.g., GPQA) and math settings.
    }
    \label{fig:rl_perf}
    \vspace{-1\baselineskip}
\end{figure}

Training with \ourmodellarge as a verifier improves downstream performance from 34.3 to 45.2, a nearly 11 point absolute gain, with improvements coming uniformly across the six benchmarks, as shown in~\Cref{fig:rl_perf}. Notably, with 77\% fewer gradient updates\footnote{
    For the same group size, General-Reasoner-7B is trained for 700 steps with rollout batch size 768, whereas we train for 120 update steps with rollout batch size 1024.
}, our model was able to slightly beat the performance of General-Reasoner-7B~\citep{ma2025general} on several benchmarks: 38.9 vs. 38.8 on GPQA-Diamond, 38.4 vs 37.9 on OlympiadBench, and 13.9 vs. 13.8 on AIME24. This shows that using \ourmodellarge can significantly improve RL training convergence. Further, using \ourmodelsmall and \ourmodellarge bring 8.8\% and 14.1\% relative gains over typically used string matching verifiers and 4.4\% and 9.4\% over General-Verifier, which has been trained with in-training-distribution verification data. These gains appear for both natural language (MMLU-Pro, GPQA) and math domains, showing that \ourmodel can verify complex outputs across multiple challenging domains.

\begin{wrapfigure}{r}{0.47\textwidth}
    \vspace{-1\baselineskip}
    \centering
    \includegraphics[width=0.41\textwidth]{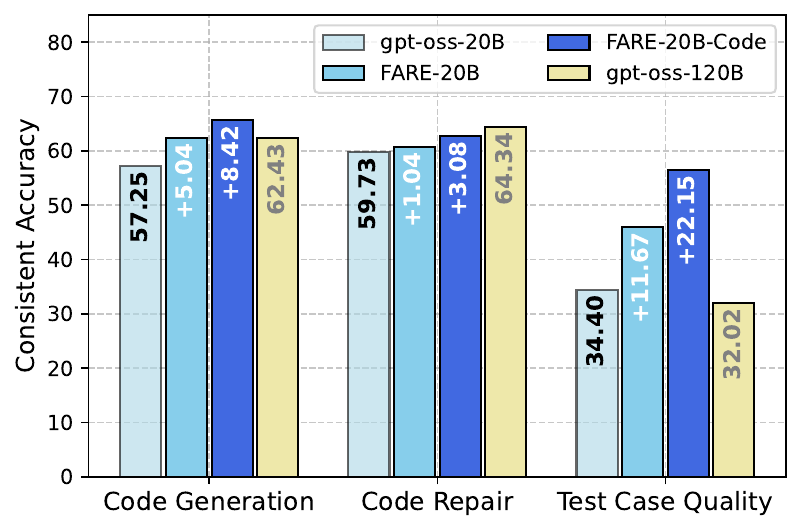}
    \caption{Continual training of \ourmodellarge for code evaluation with only 15K samples yields larger gains over gpt-oss-20B/120B.}
    \label{fig:continual}
    \vspace{-1\baselineskip}
\end{wrapfigure}

\textbf{Initialization for domain-specific continual finetuning.} We continually finetune \ourmodellarge for code evaluation with one round of RS-SFT to produce \ourmodellarge-Code. We train with only 15K pairwise samples randomly chosen from AceCoder~\citep{zeng2025acecoder}. For evaluation, we use the recently released CodingJudgeBench~\citep{jiang2025codejudgebench}, a pairwise benchmark covering code generation, code repair, and test-case quality evaluation tasks.~\Cref{fig:continual}, which reports consistent accuracy across the three splits from gpt-oss-20B, \ourmodellarge, \ourmodellarge-Code, and gpt-oss-120B for reference, shows that the first two splits are relatively easy, whereas test case evaluation is extremely difficult: gpt-oss-120B achieves only 32\%. On the last task, \ourmodellarge improves over gpt-oss-20B by 11.67 absolute points, highlighting the benefits of large-scale evaluation training. Specialized continual training on coding tasks brings an additional 10.48 absolute point improvement, with \ourmodellarge-Code outperforming even gpt-oss-120B on average. In all, \ourmodel can be readily adapted for specific applications with a small amount of domain-specific data.
\section{Conclusion}
Using a curated multi-task, multi-domain training mix and RS-SFT, we train \ourmodel, a family of high performing and well-rounded automatic evaluators. \ourmodelsmall challenges larger specialized evaluators and \ourmodellarge sets a new standard for locally hosted evaluators. Our evaluations include 7 challenging benchmarks and 3 practical downstream settings where we show that \ourmodel are (1) effective reward models at inference-time, (2) effective verifiers for GRPO training, and (3) strong initializations for continual, domain-specific finetuning.

\bibliography{iclr2026_conference}

@article{mao2024champ,
  title={CHAMP: A Competition-level Dataset for Fine-Grained Analyses of LLMs' Mathematical Reasoning Capabilities},
  author={Mao, Yujun and Kim, Yoon and Zhou, Yilun},
  journal={arXiv preprint arXiv:2401.06961},
  year={2024}
}

@inproceedings{ross2011reduction,
  title={A reduction of imitation learning and structured prediction to no-regret online learning},
  author={Ross, St{\'e}phane and Gordon, Geoffrey and Bagnell, Drew},
  booktitle={Proceedings of the fourteenth international conference on artificial intelligence and statistics},
  pages={627--635},
  year={2011},
  organization={JMLR Workshop and Conference Proceedings}
}

@article{cobbe2021training,
  title={Training verifiers to solve math word problems},
  author={Cobbe, Karl and Kosaraju, Vineet and Bavarian, Mohammad and Chen, Mark and Jun, Heewoo and Kaiser, Lukasz and Plappert, Matthias and Tworek, Jerry and Hilton, Jacob and Nakano, Reiichiro and others},
  journal={arXiv preprint arXiv:2110.14168},
  year={2021}
}

@article{luo2024improve,
  title={Improve Mathematical Reasoning in Language Models by Automated Process Supervision},
  author={Luo, Liangchen and Liu, Yinxiao and Liu, Rosanne and Phatale, Samrat and Lara, Harsh and Li, Yunxuan and Shu, Lei and Zhu, Yun and Meng, Lei and Sun, Jiao and others},
  journal={arXiv preprint arXiv:2406.06592},
  year={2024}
}

@article{wang2023math,
  title={Math-shepherd: A label-free step-by-step verifier for llms in mathematical reasoning},
  author={Wang, Peiyi and Li, Lei and Shao, Zhihong and Xu, RX and Dai, Damai and Li, Yifei and Chen, Deli and Wu, Y and Sui, Zhifang},
  journal={arXiv preprint arXiv:2312.08935},
  year={2023}
}

@article{lightman2023let,
  title={Let's verify step by step},
  author={Lightman, Hunter and Kosaraju, Vineet and Burda, Yura and Edwards, Harri and Baker, Bowen and Lee, Teddy and Leike, Jan and Schulman, John and Sutskever, Ilya and Cobbe, Karl},
  journal={arXiv preprint arXiv:2305.20050},
  year={2023}
}

@article{wang2022self,
  title={Self-consistency improves chain of thought reasoning in language models},
  author={Wang, Xuezhi and Wei, Jason and Schuurmans, Dale and Le, Quoc and Chi, Ed and Narang, Sharan and Chowdhery, Aakanksha and Zhou, Denny},
  journal={arXiv preprint arXiv:2203.11171},
  year={2022}
}

@article{zheng2023judging,
  title={Judging llm-as-a-judge with mt-bench and chatbot arena},
  author={Zheng, Lianmin and Chiang, Wei-Lin and Sheng, Ying and Zhuang, Siyuan and Wu, Zhanghao and Zhuang, Yonghao and Lin, Zi and Li, Zhuohan and Li, Dacheng and Xing, Eric and others},
  journal={Advances in Neural Information Processing Systems},
  volume={36},
  pages={46595--46623},
  year={2023}
}

@article{hendrycks2021measuring,
  title={Measuring mathematical problem solving with the math dataset},
  author={Hendrycks, Dan and Burns, Collin and Kadavath, Saurav and Arora, Akul and Basart, Steven and Tang, Eric and Song, Dawn and Steinhardt, Jacob},
  journal={arXiv preprint arXiv:2103.03874},
  year={2021}
}

@article{zeng2023evaluating,
  title={Evaluating large language models at evaluating instruction following},
  author={Zeng, Zhiyuan and Yu, Jiatong and Gao, Tianyu and Meng, Yu and Goyal, Tanya and Chen, Danqi},
  journal={arXiv preprint arXiv:2310.07641},
  year={2023}
}

@article{zheng2024processbench,
  title={Processbench: Identifying process errors in mathematical reasoning},
  author={Zheng, Chujie and Zhang, Zhenru and Zhang, Beichen and Lin, Runji and Lu, Keming and Yu, Bowen and Liu, Dayiheng and Zhou, Jingren and Lin, Junyang},
  journal={arXiv preprint arXiv:2412.06559},
  year={2024}
}

@article{tan2024judgebench,
  title={Judgebench: A benchmark for evaluating llm-based judges},
  author={Tan, Sijun and Zhuang, Siyuan and Montgomery, Kyle and Tang, William Y and Cuadron, Alejandro and Wang, Chenguang and Popa, Raluca Ada and Stoica, Ion},
  journal={arXiv preprint arXiv:2410.12784},
  year={2024}
}

@article{frick2024evaluate,
  title={How to Evaluate Reward Models for RLHF},
  author={Frick, Evan and Li, Tianle and Chen, Connor and Chiang, Wei-Lin and Angelopoulos, Anastasios N and Jiao, Jiantao and Zhu, Banghua and Gonzalez, Joseph E and Stoica, Ion},
  journal={arXiv preprint arXiv:2410.14872},
  year={2024}
}

@article{wang2024self,
  title={Self-Taught Evaluators},
  author={Wang, Tianlu and Kulikov, Ilia and Golovneva, Olga and Yu, Ping and Yuan, Weizhe and Dwivedi-Yu, Jane and Pang, Richard Yuanzhe and Fazel-Zarandi, Maryam and Weston, Jason and Li, Xian},
  journal={arXiv preprint arXiv:2408.02666},
  year={2024}
}

@article{kim2024prometheus,
  title={Prometheus 2: An open source language model specialized in evaluating other language models},
  author={Kim, Seungone and Suk, Juyoung and Longpre, Shayne and Lin, Bill Yuchen and Shin, Jamin and Welleck, Sean and Neubig, Graham and Lee, Moontae and Lee, Kyungjae and Seo, Minjoon},
  journal={arXiv preprint arXiv:2405.01535},
  year={2024}
}

@article{kim2024biggen,
  title={The BiGGen Bench: A Principled Benchmark for Fine-grained Evaluation of Language Models with Language Models},
  author={Kim, Seungone and Suk, Juyoung and Cho, Ji Yong and Longpre, Shayne and Kim, Chaeeun and Yoon, Dongkeun and Son, Guijin and Cho, Yejin and Shafayat, Sheikh and Baek, Jinheon and others},
  journal={arXiv preprint arXiv:2406.05761},
  year={2024}
}

@inproceedings{kim2023prometheus,
  title={Prometheus: Inducing fine-grained evaluation capability in language models},
  author={Kim, Seungone and Shin, Jamin and Cho, Yejin and Jang, Joel and Longpre, Shayne and Lee, Hwaran and Yun, Sangdoo and Shin, Seongjin and Kim, Sungdong and Thorne, James and others},
  booktitle={The Twelfth International Conference on Learning Representations},
  year={2023}
}

@article{li2023generative,
  title={Generative judge for evaluating alignment},
  author={Li, Junlong and Sun, Shichao and Yuan, Weizhe and Fan, Run-Ze and Zhao, Hai and Liu, Pengfei},
  journal={arXiv preprint arXiv:2310.05470},
  year={2023}
}

@article{vu2024foundational,
  title={Foundational autoraters: Taming large language models for better automatic evaluation},
  author={Vu, Tu and Krishna, Kalpesh and Alzubi, Salaheddin and Tar, Chris and Faruqui, Manaal and Sung, Yun-Hsuan},
  journal={arXiv preprint arXiv:2407.10817},
  year={2024}
}

@article{hu2024themis,
  title={Themis: A Reference-free NLG Evaluation Language Model with Flexibility and Interpretability},
  author={Hu, Xinyu and Lin, Li and Gao, Mingqi and Yin, Xunjian and Wan, Xiaojun},
  journal={arXiv preprint arXiv:2406.18365},
  year={2024}
}

@article{ye2024beyond,
  title={Beyond Scalar Reward Model: Learning Generative Judge from Preference Data},
  author={Ye, Ziyi and Li, Xiangsheng and Li, Qiuchi and Ai, Qingyao and Zhou, Yujia and Shen, Wei and Yan, Dong and Liu, Yiqun},
  journal={arXiv preprint arXiv:2410.03742},
  year={2024}
}

@article{wang2024direct,
  title={Direct judgement preference optimization},
  author={Wang, Peifeng and Xu, Austin and Zhou, Yilun and Xiong, Caiming and Joty, Shafiq},
  journal={arXiv preprint arXiv:2409.14664},
  year={2024}
}

@article{saad2024lmunit,
  title={LMUnit: Fine-grained Evaluation with Natural Language Unit Tests},
  author={Saad-Falcon, Jon and Vivek, Rajan and Berrios, William and Naik, Nandita Shankar and Franklin, Matija and Vidgen, Bertie and Singh, Amanpreet and Kiela, Douwe and Mehri, Shikib},
  journal={arXiv preprint arXiv:2412.13091},
  year={2024}
}

@article{zhuo2024bigcodebench,
  title={Bigcodebench: Benchmarking code generation with diverse function calls and complex instructions},
  author={Zhuo, Terry Yue and Vu, Minh Chien and Chim, Jenny and Hu, Han and Yu, Wenhao and Widyasari, Ratnadira and Yusuf, Imam Nur Bani and Zhan, Haolan and He, Junda and Paul, Indraneil and others},
  journal={arXiv preprint arXiv:2406.15877},
  year={2024}
}

@inproceedings{evalplus,
  title = {Is Your Code Generated by Chat{GPT} Really Correct? Rigorous Evaluation of Large Language Models for Code Generation},
  author = {Liu, Jiawei and Xia, Chunqiu Steven and Wang, Yuyao and Zhang, Lingming},
  booktitle = {Thirty-seventh Conference on Neural Information Processing Systems},
  year = {2023},
  url = {https://openreview.net/forum?id=1qvx610Cu7},
}

@article{zhou2023instruction,
  title={Instruction-following evaluation for large language models},
  author={Zhou, Jeffrey and Lu, Tianjian and Mishra, Swaroop and Brahma, Siddhartha and Basu, Sujoy and Luan, Yi and Zhou, Denny and Hou, Le},
  journal={arXiv preprint arXiv:2311.07911},
  year={2023}
}

@article{zhang2024generative,
  title={Generative verifiers: Reward modeling as next-token prediction},
  author={Zhang, Lunjun and Hosseini, Arian and Bansal, Hritik and Kazemi, Mehran and Kumar, Aviral and Agarwal, Rishabh},
  journal={arXiv preprint arXiv:2408.15240},
  year={2024}
}

@inproceedings{yuan2024self,
  title={Self-Rewarding Language Models},
  author={Yuan, Weizhe and Pang, Richard Yuanzhe and Cho, Kyunghyun and Li, Xian and Sukhbaatar, Sainbayar and Xu, Jing and Weston, Jason E},
  booktitle={Forty-first International Conference on Machine Learning},
  year={2024}
}

@misc{skyworkcritic2024,
  title={Skywork Critic Model Series},
  author={Shiwen, Tu and Liang, Zhao and Liu, Chris Yuhao and Zeng, Liang and Liu, Yang},
  year={2024},
  month={September},
  howpublished={\url{https://huggingface.co/Skywork}},
  url={https://huggingface.co/Skywork},
}

@article{park2024offsetbias,
  title={Offsetbias: Leveraging debiased data for tuning evaluators},
  author={Park, Junsoo and Jwa, Seungyeon and Ren, Meiying and Kim, Daeyoung and Choi, Sanghyuk},
  journal={arXiv preprint arXiv:2407.06551},
  year={2024}
}

@article{prmlessons,
  title={The Lessons of Developing Process Reward Models in Mathematical Reasoning}, 
  author={
    Zhenru Zhang and Chujie Zheng and Yangzhen Wu and Beichen Zhang and Runji Lin and Bowen Yu and Dayiheng Liu and Jingren Zhou and Junyang Lin
  },
  journal={arXiv preprint arXiv:2501.07301},
  year={2025}
}

@article{liu2024skywork,
  title={Skywork-reward: Bag of tricks for reward modeling in llms},
  author={Liu, Chris Yuhao and Zeng, Liang and Liu, Jiacai and Yan, Rui and He, Jujie and Wang, Chaojie and Yan, Shuicheng and Liu, Yang and Zhou, Yahui},
  journal={arXiv preprint arXiv:2410.18451},
  year={2024}
}

@article{dubey2024llama,
  title={The llama 3 herd of models},
  author={Dubey, Abhimanyu and Jauhri, Abhinav and Pandey, Abhinav and Kadian, Abhishek and Al-Dahle, Ahmad and Letman, Aiesha and Mathur, Akhil and Schelten, Alan and Yang, Amy and Fan, Angela and others},
  journal={arXiv preprint arXiv:2407.21783},
  year={2024}
}

@article{yang2024qwen2,
  title={Qwen2. 5 Technical Report},
  author={Yang, An and Yang, Baosong and Zhang, Beichen and Hui, Binyuan and Zheng, Bo and Yu, Bowen and Li, Chengyuan and Liu, Dayiheng and Huang, Fei and Wei, Haoran and others},
  journal={arXiv preprint arXiv:2412.15115},
  year={2024}
}

@article{dubois2024length,
  title={Length-controlled alpacaeval: A simple way to debias automatic evaluators},
  author={Dubois, Yann and Galambosi, Bal{\'a}zs and Liang, Percy and Hashimoto, Tatsunori B},
  journal={arXiv preprint arXiv:2404.04475},
  year={2024}
}

@article{liu2024rm,
  title={RM-bench: Benchmarking reward models of language models with subtlety and style},
  author={Liu, Yantao and Yao, Zijun and Min, Rui and Cao, Yixin and Hou, Lei and Li, Juanzi},
  journal={arXiv preprint arXiv:2410.16184},
  year={2024}
}

@article{ke2025survey,
  title={A Survey of Frontiers in LLM Reasoning: Inference Scaling, Learning to Reason, and Agentic Systems},
  author={Ke, Zixuan and Jiao, Fangkai and Ming, Yifei and Nguyen, Xuan-Phi and Xu, Austin and Long, Do Xuan and Li, Minzhi and Qin, Chengwei and Wang, Peifeng and Savarese, Silvio and others},
  journal={arXiv preprint arXiv:2504.09037},
  year={2025}
}

@article{zhou2025evaluating,
  title={Evaluating Judges as Evaluators: The JETTS Benchmark of LLM-as-Judges as Test-Time Scaling Evaluators},
  author={Zhou, Yilun and Xu, Austin and Wang, Peifeng and Xiong, Caiming and Joty, Shafiq},
  journal={arXiv preprint arXiv:2504.15253},
  year={2025}
}

@article{lambert2024t,
  title={T$\backslash$" ulu 3: Pushing frontiers in open language model post-training},
  author={Lambert, Nathan and Morrison, Jacob and Pyatkin, Valentina and Huang, Shengyi and Ivison, Hamish and Brahman, Faeze and Miranda, Lester James V and Liu, Alisa and Dziri, Nouha and Lyu, Shane and others},
  journal={arXiv preprint arXiv:2411.15124},
  year={2024}
}

@article{wang2023large,
  title={Large language models are not fair evaluators},
  author={Wang, Peiyi and Li, Lei and Chen, Liang and Cai, Zefan and Zhu, Dawei and Lin, Binghuai and Cao, Yunbo and Liu, Qi and Liu, Tianyu and Sui, Zhifang},
  journal={arXiv preprint arXiv:2305.17926},
  year={2023}
}

@article{xu2025does,
  title={Does context matter? contextualjudgebench for evaluating llm-based judges in contextual settings},
  author={Xu, Austin and Bansal, Srijan and Ming, Yifei and Yavuz, Semih and Joty, Shafiq},
  journal={arXiv preprint arXiv:2503.15620},
  year={2025}
}

@article{han2022folio,
  title={Folio: Natural language reasoning with first-order logic},
  author={Han, Simeng and Schoelkopf, Hailey and Zhao, Yilun and Qi, Zhenting and Riddell, Martin and Zhou, Wenfei and Coady, James and Peng, David and Qiao, Yujie and Benson, Luke and others},
  journal={arXiv preprint arXiv:2209.00840},
  year={2022}
}

@article{geva2021did,
  title={Did aristotle use a laptop? a question answering benchmark with implicit reasoning strategies},
  author={Geva, Mor and Khashabi, Daniel and Segal, Elad and Khot, Tushar and Roth, Dan and Berant, Jonathan},
  journal={Transactions of the Association for Computational Linguistics},
  volume={9},
  pages={346--361},
  year={2021},
  publisher={MIT Press One Rogers Street, Cambridge, MA 02142-1209, USA journals-info~…}
}

@article{yu2020reclor,
  title={Reclor: A reading comprehension dataset requiring logical reasoning},
  author={Yu, Weihao and Jiang, Zihang and Dong, Yanfei and Feng, Jiashi},
  journal={arXiv preprint arXiv:2002.04326},
  year={2020}
}

@article{cao2024compassjudger,
  title={Compassjudger-1: All-in-one judge model helps model evaluation and evolution},
  author={Cao, Maosong and Lam, Alexander and Duan, Haodong and Liu, Hongwei and Zhang, Songyang and Chen, Kai},
  journal={arXiv preprint arXiv:2410.16256},
  year={2024}
}

@article{saha2025learning,
  title={Learning to Plan \& Reason for Evaluation with Thinking-LLM-as-a-Judge},
  author={Saha, Swarnadeep and Li, Xian and Ghazvininejad, Marjan and Weston, Jason and Wang, Tianlu},
  journal={arXiv preprint arXiv:2501.18099},
  year={2025}
}

@article{schulman2017proximal,
  title={Proximal policy optimization algorithms},
  author={Schulman, John and Wolski, Filip and Dhariwal, Prafulla and Radford, Alec and Klimov, Oleg},
  journal={arXiv preprint arXiv:1707.06347},
  year={2017}
}

@article{yu2025dapo,
  title={Dapo: An open-source llm reinforcement learning system at scale},
  author={Yu, Qiying and Zhang, Zheng and Zhu, Ruofei and Yuan, Yufeng and Zuo, Xiaochen and Yue, Yu and Fan, Tiantian and Liu, Gaohong and Liu, Lingjun and Liu, Xin and others},
  journal={arXiv preprint arXiv:2503.14476},
  year={2025}
}

@article{chen2025judgelrm,
  title={Judgelrm: Large reasoning models as a judge},
  author={Chen, Nuo and Hu, Zhiyuan and Zou, Qingyun and Wu, Jiaying and Wang, Qian and Hooi, Bryan and He, Bingsheng},
  journal={arXiv preprint arXiv:2504.00050},
  year={2025}
}

@article{hu2024openrlhf,
  title={Openrlhf: An easy-to-use, scalable and high-performance rlhf framework},
  author={Hu, Jian and Wu, Xibin and Zhu, Zilin and Wang, Weixun and Zhang, Dehao and Cao, Yu and others},
  journal={arXiv preprint arXiv:2405.11143},
  year={2024}
}

@article{hu2024automated,
  title={Automated design of agentic systems},
  author={Hu, Shengran and Lu, Cong and Clune, Jeff},
  journal={arXiv preprint arXiv:2408.08435},
  year={2024}
}

@article{mahan2024generative,
  title={Generative reward models},
  author={Mahan, Dakota and Van Phung, Duy and Rafailov, Rafael and Blagden, Chase and Lile, Nathan and Castricato, Louis and Fr{\"a}nken, Jan-Philipp and Finn, Chelsea and Albalak, Alon},
  journal={arXiv preprint arXiv:2410.12832},
  year={2024}
}

@misc{chen2025rmr1rewardmodelingreasoning,
      title={RM-R1: Reward Modeling as Reasoning}, 
      author={Xiusi Chen and Gaotang Li and Ziqi Wang and Bowen Jin and Cheng Qian and Yu Wang and Hongru Wang and Yu Zhang and Denghui Zhang and Tong Zhang and Hanghang Tong and Heng Ji},
      year={2025},
      eprint={2505.02387},
      archivePrefix={arXiv},
      primaryClass={cs.CL},
      url={https://arxiv.org/abs/2505.02387}, 
}

@article{liu2025inference,
  title={Inference-time scaling for generalist reward modeling},
  author={Liu, Zijun and Wang, Peiyi and Xu, Runxin and Ma, Shirong and Ruan, Chong and Li, Peng and Liu, Yang and Wu, Yu},
  journal={arXiv preprint arXiv:2504.02495},
  year={2025}
}

@article{stiennon2020learning,
  title={Learning to summarize with human feedback},
  author={Stiennon, Nisan and Ouyang, Long and Wu, Jeffrey and Ziegler, Daniel and Lowe, Ryan and Voss, Chelsea and Radford, Alec and Amodei, Dario and Christiano, Paul F},
  journal={Advances in neural information processing systems},
  volume={33},
  pages={3008--3021},
  year={2020}
}

@article{li2024crowdsourced,
  title={From crowdsourced data to high-quality benchmarks: Arena-hard and benchbuilder pipeline},
  author={Li, Tianle and Chiang, Wei-Lin and Frick, Evan and Dunlap, Lisa and Wu, Tianhao and Zhu, Banghua and Gonzalez, Joseph E and Stoica, Ion},
  journal={arXiv preprint arXiv:2406.11939},
  year={2024}
}

@inproceedings{rein2024gpqa,
  title={Gpqa: A graduate-level google-proof q\&a benchmark},
  author={Rein, David and Hou, Betty Li and Stickland, Asa Cooper and Petty, Jackson and Pang, Richard Yuanzhe and Dirani, Julien and Michael, Julian and Bowman, Samuel R},
  booktitle={First Conference on Language Modeling},
  year={2024}
}

@article{he2024olympiadbench,
  title={Olympiadbench: A challenging benchmark for promoting agi with olympiad-level bilingual multimodal scientific problems},
  author={He, Chaoqun and Luo, Renjie and Bai, Yuzhuo and Hu, Shengding and Thai, Zhen Leng and Shen, Junhao and Hu, Jinyi and Han, Xu and Huang, Yujie and Zhang, Yuxiang and others},
  journal={arXiv preprint arXiv:2402.14008},
  year={2024}
}

@inproceedings{wang2024mmlu,
  title={Mmlu-pro: A more robust and challenging multi-task language understanding benchmark},
  author={Wang, Yubo and Ma, Xueguang and Zhang, Ge and Ni, Yuansheng and Chandra, Abhranil and Guo, Shiguang and Ren, Weiming and Arulraj, Aaran and He, Xuan and Jiang, Ziyan and others},
  booktitle={The Thirty-eight Conference on Neural Information Processing Systems Datasets and Benchmarks Track},
  year={2024}
}

@article{yu2025improve,
  title={Improve LLM-as-a-Judge Ability as a General Ability},
  author={Yu, Jiachen and Sun, Shaoning and Hu, Xiaohui and Yan, Jiaxu and Yu, Kaidong and Li, Xuelong},
  journal={arXiv preprint arXiv:2502.11689},
  year={2025}
}

@article{khalifa2025process,
  title={Process reward models that think},
  author={Khalifa, Muhammad and Agarwal, Rishabh and Logeswaran, Lajanugen and Kim, Jaekyeom and Peng, Hao and Lee, Moontae and Lee, Honglak and Wang, Lu},
  journal={arXiv preprint arXiv:2504.16828},
  year={2025}
}

@article{mcaleese2024llm,
  title={Llm critics help catch llm bugs},
  author={McAleese, Nat and Pokorny, Rai Michael and Uribe, Juan Felipe Ceron and Nitishinskaya, Evgenia and Trebacz, Maja and Leike, Jan},
  journal={arXiv preprint arXiv:2407.00215},
  year={2024}
}

@article{cemri2025multi,
  title={Why Do Multi-Agent LLM Systems Fail?},
  author={Cemri, Mert and Pan, Melissa Z and Yang, Shuyi and Agrawal, Lakshya A and Chopra, Bhavya and Tiwari, Rishabh and Keutzer, Kurt and Parameswaran, Aditya and Klein, Dan and Ramchandran, Kannan and others},
  journal={arXiv preprint arXiv:2503.13657},
  year={2025}
}

@article{chan2025j1,
  title={J1: Exploring Simple Test-Time Scaling for LLM-as-a-Judge},
  author={Chan, Chi-Min and Xu, Chunpu and Ji, Jiaming and Ye, Zhen and Wen, Pengcheng and Jiang, Chunyang and Yang, Yaodong and Xue, Wei and Han, Sirui and Guo, Yike},
  journal={arXiv preprint arXiv:2505.11875},
  year={2025}
}

@article{whitehouse2025j1,
  title={J1: Incentivizing thinking in llm-as-a-judge via reinforcement learning},
  author={Whitehouse, Chenxi and Wang, Tianlu and Yu, Ping and Li, Xian and Weston, Jason and Kulikov, Ilia and Saha, Swarnadeep},
  journal={arXiv preprint arXiv:2505.10320},
  year={2025}
}

@article{xu2025j4r,
  title={J4R: Learning to Judge with Equivalent Initial State Group Relative Policy Optimization},
  author={Xu, Austin and Zhou, Yilun and Nguyen, Xuan-Phi and Xiong, Caiming and Joty, Shafiq},
  journal={arXiv preprint arXiv:2505.13346},
  year={2025}
}

@article{guha2025openthoughts,
  title={OpenThoughts: Data Recipes for Reasoning Models},
  author={Guha, Etash and Marten, Ryan and Keh, Sedrick and Raoof, Negin and Smyrnis, Georgios and Bansal, Hritik and Nezhurina, Marianna and Mercat, Jean and Vu, Trung and Sprague, Zayne and others},
  journal={arXiv preprint arXiv:2506.04178},
  year={2025}
}

@article{liu2025prorl,
  title={Prorl: Prolonged reinforcement learning expands reasoning boundaries in large language models},
  author={Liu, Mingjie and Diao, Shizhe and Lu, Ximing and Hu, Jian and Dong, Xin and Choi, Yejin and Kautz, Jan and Dong, Yi},
  journal={arXiv preprint arXiv:2505.24864},
  year={2025}
}

@article{xiong2025minimalist,
  title={A minimalist approach to llm reasoning: from rejection sampling to reinforce},
  author={Xiong, Wei and Yao, Jiarui and Xu, Yuhui and Pang, Bo and Wang, Lei and Sahoo, Doyen and Li, Junnan and Jiang, Nan and Zhang, Tong and Xiong, Caiming and others},
  journal={arXiv preprint arXiv:2504.11343},
  year={2025}
}

@article{ross2025when2call,
  title={When2Call: When (not) to Call Tools},
  author={Ross, Hayley and Mahabaleshwarkar, Ameya Sunil and Suhara, Yoshi},
  journal={arXiv preprint arXiv:2504.18851},
  year={2025}
}

@article{panickssery2024llm,
  title={Llm evaluators recognize and favor their own generations},
  author={Panickssery, Arjun and Bowman, Samuel and Feng, Shi},
  journal={Advances in Neural Information Processing Systems},
  volume={37},
  pages={68772--68802},
  year={2024}
}

@inproceedings{wang2023chatgpt,
  title={Is ChatGPT a Good NLG Evaluator? A Preliminary Study},
  author={Wang, Jiaan and Liang, Yunlong and Meng, Fandong and Sun, Zengkui and Shi, Haoxiang and Li, Zhixu and Xu, Jinan and Qu, Jianfeng and Zhou, Jie},
  booktitle={Proceedings of EMNLP Workshop},
  pages={1},
  year={2023}
}

@inproceedings{liu2023g,
  title={G-Eval: NLG Evaluation using Gpt-4 with Better Human Alignment},
  author={Liu, Yang and Iter, Dan and Xu, Yichong and Wang, Shuohang and Xu, Ruochen and Zhu, Chenguang},
  booktitle={Proceedings of the 2023 Conference on Empirical Methods in Natural Language Processing},
  pages={2511--2522},
  year={2023}
}

@inproceedings{fu2024gptscore,
  title={GPTScore: Evaluate as You Desire},
  author={Fu, Jinlan and Ng, See Kiong and Jiang, Zhengbao and Liu, Pengfei},
  booktitle={Proceedings of the 2024 Conference of the North American Chapter of the Association for Computational Linguistics: Human Language Technologies (Volume 1: Long Papers)},
  pages={6556--6576},
  year={2024}
}

@inproceedings{chiang2023can,
  title={Can Large Language Models Be an Alternative to Human Evaluations?},
  author={Chiang, Cheng-Han and Lee, Hung-Yi},
  booktitle={Proceedings of the 61st Annual Meeting of the Association for Computational Linguistics (Volume 1: Long Papers)},
  pages={15607--15631},
  year={2023}
}

@article{yan2025verifybench,
  title={VerifyBench: Benchmarking Reference-based Reward Systems for Large Language Models},
  author={Yan, Yuchen and Jiang, Jin and Ren, Zhenbang and Li, Yijun and Cai, Xudong and Liu, Yang and Xu, Xin and Zhang, Mengdi and Shao, Jian and Shen, Yongliang and others},
  journal={arXiv preprint arXiv:2505.15801},
  year={2025}
}

@article{zelikman2022star,
  title={Star: Bootstrapping reasoning with reasoning},
  author={Zelikman, Eric and Wu, Yuhuai and Mu, Jesse and Goodman, Noah},
  journal={Advances in Neural Information Processing Systems},
  volume={35},
  pages={15476--15488},
  year={2022}
}

@article{dong2023raft,
  title={Raft: Reward ranked finetuning for generative foundation model alignment},
  author={Dong, Hanze and Xiong, Wei and Goyal, Deepanshu and Zhang, Yihan and Chow, Winnie and Pan, Rui and Diao, Shizhe and Zhang, Jipeng and Shum, Kashun and Zhang, Tong},
  journal={arXiv preprint arXiv:2304.06767},
  year={2023}
}

@article{touvron2023llama,
  title={Llama 2: Open foundation and fine-tuned chat models},
  author={Touvron, Hugo and Martin, Louis and Stone, Kevin and Albert, Peter and Almahairi, Amjad and Babaei, Yasmine and Bashlykov, Nikolay and Batra, Soumya and Bhargava, Prajjwal and Bhosale, Shruti and others},
  journal={arXiv preprint arXiv:2307.09288},
  year={2023}
}

@article{ma2025general,
  title={General-reasoner: Advancing llm reasoning across all domains},
  author={Ma, Xueguang and Liu, Qian and Jiang, Dongfu and Zhang, Ge and Ma, Zejun and Chen, Wenhu},
  journal={arXiv preprint arXiv:2505.14652},
  year={2025}
}

@article{lewkowycz2022solving,
  title={Solving quantitative reasoning problems with language models},
  author={Lewkowycz, Aitor and Andreassen, Anders and Dohan, David and Dyer, Ethan and Michalewski, Henryk and Ramasesh, Vinay and Slone, Ambrose and Anil, Cem and Schlag, Imanol and Gutman-Solo, Theo and others},
  journal={Advances in neural information processing systems},
  volume={35},
  pages={3843--3857},
  year={2022}
}

@article{wu2024meta,
  title={Meta-rewarding language models: Self-improving alignment with llm-as-a-meta-judge},
  author={Wu, Tianhao and Yuan, Weizhe and Golovneva, Olga and Xu, Jing and Tian, Yuandong and Jiao, Jiantao and Weston, Jason and Sukhbaatar, Sainbayar},
  journal={arXiv preprint arXiv:2407.19594},
  year={2024}
}

@article{team2025kimi,
  title={Kimi k2: Open agentic intelligence},
  author={{Team Kimi} and Bai, Yifan and Bao, Yiping and Chen, Guanduo and Chen, Jiahao and Chen, Ningxin and Chen, Ruijue and Chen, Yanru and Chen, Yuankun and Chen, Yutian and others},
  journal={arXiv preprint arXiv:2507.20534},
  year={2025}
}

@article{wu2025sailing,
  title={Sailing ai by the stars: A survey of learning from rewards in post-training and test-time scaling of large language models},
  author={Wu, Xiaobao},
  journal={arXiv preprint arXiv:2505.02686},
  year={2025}
}

@article{ye2023flask,
  title={Flask: Fine-grained language model evaluation based on alignment skill sets},
  author={Ye, Seonghyeon and Kim, Doyoung and Kim, Sungdong and Hwang, Hyeonbin and Kim, Seungone and Jo, Yongrae and Thorne, James and Kim, Juho and Seo, Minjoon},
  journal={arXiv preprint arXiv:2307.10928},
  year={2023}
}

@article{dsr1_guo2025deepseek,
  title={Deepseek-r1: Incentivizing reasoning capability in llms via reinforcement learning},
  author={Guo, Daya and Yang, Dejian and Zhang, Haowei and Song, Junxiao and Zhang, Ruoyu and Xu, Runxin and Zhu, Qihao and Ma, Shirong and Wang, Peiyi and Bi, Xiao and others},
  journal={arXiv preprint arXiv:2501.12948},
  year={2025}
}

@article{yang2025qwen3,
  title={Qwen3 technical report},
  author={Yang, An and Li, Anfeng and Yang, Baosong and Zhang, Beichen and Hui, Binyuan and Zheng, Bo and Yu, Bowen and Gao, Chang and Huang, Chengen and Lv, Chenxu and others},
  journal={arXiv preprint arXiv:2505.09388},
  year={2025}
}

@article{instructgpt_ouyang2022training,
  title={Training language models to follow instructions with human feedback},
  author={Ouyang, Long and Wu, Jeffrey and Jiang, Xu and Almeida, Diogo and Wainwright, Carroll and Mishkin, Pamela and Zhang, Chong and Agarwal, Sandhini and Slama, Katarina and Ray, Alex and others},
  journal={Advances in Neural Information Processing Systems},
  volume={35},
  pages={27730--27744},
  year={2022}
}

@article{dpo_rafailov2023direct,
  title={Direct preference optimization: Your language model is secretly a reward model},
  author={Rafailov, Rafael and Sharma, Archit and Mitchell, Eric and Ermon, Stefano and Manning, Christopher D and Finn, Chelsea},
  journal={arXiv preprint arXiv:2305.18290},
  year={2023}
}

@misc{deepscaler2025,
  title={DeepScaleR: Surpassing O1-Preview with a 1.5B Model by Scaling RL},
  author={Michael Luo and Sijun Tan and Justin Wong and Xiaoxiang Shi and William Tang and Manan Roongta and Colin Cai and Jeffrey Luo and Tianjun Zhang and Erran Li and Raluca Ada Popa and Ion Stoica},
  year={2025},
  howpublished={\url{https://pretty-radio-b75.notion.site/DeepScaleR-Surpassing-O1-Preview-with-a-1-5B-Model-by-Scaling-RL-19681902c1468005bed8ca303013a4e2}},
  note={Notion Blog},
}

@article{duan2025efficient,
  title={Efficient process reward model training via active learning},
  author={Duan, Keyu and Liu, Zichen and Mao, Xin and Pang, Tianyu and Chen, Changyu and Chen, Qiguang and Shieh, Michael Qizhe and Dou, Longxu},
  journal={arXiv preprint arXiv:2504.10559},
  year={2025}
}

@article{ji2023beavertails,
  title={Beavertails: Towards improved safety alignment of llm via a human-preference dataset},
  author={Ji, Jiaming and Liu, Mickel and Dai, Josef and Pan, Xuehai and Zhang, Chi and Bian, Ce and Chen, Boyuan and Sun, Ruiyang and Wang, Yizhou and Yang, Yaodong},
  journal={Advances in Neural Information Processing Systems},
  volume={36},
  pages={24678--24704},
  year={2023}
}

@article{kamoi2025training,
  title={Training Step-Level Reasoning Verifiers with Formal Verification Tools},
  author={Kamoi, Ryo and Zhang, Yusen and Zhang, Nan and Das, Sarkar Snigdha Sarathi and Zhang, Rui},
  journal={arXiv preprint arXiv:2505.15960},
  year={2025}
}

@article{wang2023helpsteer,
  title={Helpsteer: Multi-attribute helpfulness dataset for steerlm},
  author={Wang, Zhilin and Dong, Yi and Zeng, Jiaqi and Adams, Virginia and Sreedhar, Makesh Narsimhan and Egert, Daniel and Delalleau, Olivier and Scowcroft, Jane Polak and Kant, Neel and Swope, Aidan and others},
  journal={arXiv preprint arXiv:2311.09528},
  year={2023}
}

@article{wang2024helpsteer,
  title={Helpsteer 2: Open-source dataset for training top-performing reward models},
  author={Wang, Zhilin and Dong, Yi and Delalleau, Olivier and Zeng, Jiaqi and Shen, Gerald and Egert, Daniel and Zhang, Jimmy and Sreedhar, Makesh Narsimhan and Kuchaiev, Oleksii},
  journal={Advances in Neural Information Processing Systems},
  volume={37},
  pages={1474--1501},
  year={2024}
}

@article{wang2025helpsteer3,
  title={HelpSteer3: Human-Annotated Feedback and Edit Data to Empower Inference-Time Scaling in Open-Ended General-Domain Tasks},
  author={Wang, Zhilin and Zeng, Jiaqi and Delalleau, Olivier and Egert, Daniel and Evans, Ellie and Shin, Hoo-Chang and Soares, Felipe and Dong, Yi and Kuchaiev, Oleksii},
  journal={arXiv preprint arXiv:2503.04378},
  year={2025}
}

@article{wang2025helpsteer3pref,
  title={HelpSteer3-Preference: Open Human-Annotated Preference Data across Diverse Tasks and Languages},
  author={Wang, Zhilin and Zeng, Jiaqi and Delalleau, Olivier and Shin, Hoo-Chang and Soares, Felipe and Bukharin, Alexander and Evans, Ellie and Dong, Yi and Kuchaiev, Oleksii},
  journal={arXiv preprint arXiv:2505.11475},
  year={2025}
}

@article{bai2022training,
  title={Training a helpful and harmless assistant with reinforcement learning from human feedback},
  author={Bai, Yuntao and Jones, Andy and Ndousse, Kamal and Askell, Amanda and Chen, Anna and DasSarma, Nova and Drain, Dawn and Fort, Stanislav and Ganguli, Deep and Henighan, Tom and others},
  journal={arXiv preprint arXiv:2204.05862},
  year={2022}
}

@article{ganguli2022red,
  title={Red teaming language models to reduce harms: Methods, scaling behaviors, and lessons learned},
  author={Ganguli, Deep and Lovitt, Liane and Kernion, Jackson and Askell, Amanda and Bai, Yuntao and Kadavath, Saurav and Mann, Ben and Perez, Ethan and Schiefer, Nicholas and Ndousse, Kamal and others},
  journal={arXiv preprint arXiv:2209.07858},
  year={2022}
}

@inproceedings{chakrabarty2025can,
  title={Can ai writing be salvaged? mitigating idiosyncrasies and improving human-ai alignment in the writing process through edits},
  author={Chakrabarty, Tuhin and Laban, Philippe and Wu, Chien-Sheng},
  booktitle={Proceedings of the 2025 CHI Conference on Human Factors in Computing Systems},
  pages={1--33},
  year={2025}
}

@article{lai2024step,
  title={Step-dpo: Step-wise preference optimization for long-chain reasoning of llms},
  author={Lai, Xin and Tian, Zhuotao and Chen, Yukang and Yang, Senqiao and Peng, Xiangru and Jia, Jiaya},
  journal={arXiv preprint arXiv:2406.18629},
  year={2024}
}

@article{liu2024apigen,
  title={Apigen: Automated pipeline for generating verifiable and diverse function-calling datasets},
  author={Liu, Zuxin and Hoang, Thai and Zhang, Jianguo and Zhu, Ming and Lan, Tian and Tan, Juntao and Yao, Weiran and Liu, Zhiwei and Feng, Yihao and RN, Rithesh and others},
  journal={Advances in Neural Information Processing Systems},
  volume={37},
  pages={54463--54482},
  year={2024}
}

@article{reddy2025swerank,
  title={SweRank: Software Issue Localization with Code Ranking},
  author={Reddy, Revanth Gangi and Suresh, Tarun and Doo, JaeHyeok and Liu, Ye and Nguyen, Xuan Phi and Zhou, Yingbo and Yavuz, Semih and Xiong, Caiming and Ji, Heng and Joty, Shafiq},
  journal={arXiv preprint arXiv:2505.07849},
  year={2025}
}

@article{liu2025synlogic,
  title={SynLogic: Synthesizing Verifiable Reasoning Data at Scale for Learning Logical Reasoning and Beyond},
  author={Liu, Junteng and Fan, Yuanxiang and Jiang, Zhuo and Ding, Han and Hu, Yongyi and Zhang, Chi and Shi, Yiqi and Weng, Shitong and Chen, Aili and Chen, Shiqi and others},
  journal={arXiv preprint arXiv:2505.19641},
  year={2025}
}

@article{pan2024training,
  title={Training software engineering agents and verifiers with swe-gym},
  author={Pan, Jiayi and Wang, Xingyao and Neubig, Graham and Jaitly, Navdeep and Ji, Heng and Suhr, Alane and Zhang, Yizhe},
  journal={arXiv preprint arXiv:2412.21139},
  year={2024}
}

@article{agarwal2025gpt,
  title={gpt-oss-120b \& gpt-oss-20b model card},
  author={Agarwal, Sandhini and Ahmad, Lama and Ai, Jason and Altman, Sam and Applebaum, Andy and Arbus, Edwin and Arora, Rahul K and Bai, Yu and Baker, Bowen and Bao, Haiming and others},
  journal={arXiv preprint arXiv:2508.10925},
  year={2025}
}

@article{jiang2025codejudgebench,
  title={CodeJudgeBench: Benchmarking LLM-as-a-Judge for Coding Tasks},
  author={Jiang, Hongchao and Chen, Yiming and Cao, Yushi and Lee, Hung-yi and Tan, Robby T},
  journal={arXiv preprint arXiv:2507.10535},
  year={2025}
}

@article{zeng2025acecoder,
  title={Acecoder: Acing coder rl via automated test-case synthesis},
  author={Zeng, Huaye and Jiang, Dongfu and Wang, Haozhe and Nie, Ping and Chen, Xiaotong and Chen, Wenhu},
  journal={arXiv preprint arXiv:2502.01718},
  year={2025}
}

@article{alexandru2025atla,
  title={Atla selene mini: A general purpose evaluation model},
  author={Alexandru, Andrei and Calvi, Antonia and Broomfield, Henry and Golden, Jackson and Dai, Kyle and Leys, Mathias and Burger, Maurice and Bartolo, Max and Engeler, Roman and Pisupati, Sashank and others},
  journal={arXiv preprint arXiv:2501.17195},
  year={2025}
}

@misc{he_2024_16998085,
  author       = {He, Jujie and
                  Wei, Tianwen and
                  Yan, Rui and
                  Liu, Jiacai and
                  Wang, Chaojie and
                  Gan, Yimeng and
                  Tu, Shiwen and
                  Liu, Chris Yuhao and
                  Zeng, Liang and
                  Wang, Xiaokun and
                  Wang, Boyang and
                  Li, Yongcong and
                  Zhang, Fuxiang and
                  Xu, Jiacheng and
                  An, Bo and
                  Liu, Yang and
                  Zhou, Yahui},
  title        = {Skywork-o1 Open Series},
  month        = nov,
  year         = 2024,
  publisher    = {Zenodo},
  version      = {1.0.0},
  doi          = {10.5281/zenodo.16998085},
  url          = {https://doi.org/10.5281/zenodo.16998085},
}

@article{zha2025rl,
  title={RL Tango: Reinforcing Generator and Verifier Together for Language Reasoning},
  author={Zha, Kaiwen and Gao, Zhengqi and Shen, Maohao and Hong, Zhang-Wei and Boning, Duane S and Katabi, Dina},
  journal={arXiv preprint arXiv:2505.15034},
  year={2025}
}

@article{xiong2025stepwiser,
  title={StepWiser: Stepwise Generative Judges for Wiser Reasoning},
  author={Xiong, Wei and Zhao, Wenting and Yuan, Weizhe and Golovneva, Olga and Zhang, Tong and Weston, Jason and Sukhbaatar, Sainbayar},
  journal={arXiv preprint arXiv:2508.19229},
  year={2025}
}

@article{sanh2021multitask,
  title={Multitask prompted training enables zero-shot task generalization},
  author={Sanh, Victor and Webson, Albert and Raffel, Colin and Bach, Stephen H and Sutawika, Lintang and Alyafeai, Zaid and Chaffin, Antoine and Stiegler, Arnaud and Scao, Teven Le and Raja, Arun and others},
  journal={arXiv preprint arXiv:2110.08207},
  year={2021}
}

@article{raffel2020exploring,
  title={Exploring the limits of transfer learning with a unified text-to-text transformer},
  author={Raffel, Colin and Shazeer, Noam and Roberts, Adam and Lee, Katherine and Narang, Sharan and Matena, Michael and Zhou, Yanqi and Li, Wei and Liu, Peter J},
  journal={Journal of machine learning research},
  volume={21},
  number={140},
  pages={1--67},
  year={2020}
}

@article{wei2021finetuned,
  title={Finetuned language models are zero-shot learners},
  author={Wei, Jason and Bosma, Maarten and Zhao, Vincent Y and Guu, Kelvin and Yu, Adams Wei and Lester, Brian and Du, Nan and Dai, Andrew M and Le, Quoc V},
  journal={arXiv preprint arXiv:2109.01652},
  year={2021}
}

@article{zhao2025genprm,
  title={Genprm: Scaling test-time compute of process reward models via generative reasoning},
  author={Zhao, Jian and Liu, Runze and Zhang, Kaiyan and Zhou, Zhimu and Gao, Junqi and Li, Dong and Lyu, Jiafei and Qian, Zhouyi and Qi, Biqing and Li, Xiu and others},
  journal={arXiv preprint arXiv:2504.00891},
  year={2025}
}

@misc{ministral8b,
    title={Un ministral, Des Ministraux},
    author={Mistral Team},
    url={https://mistral.ai/news/ministraux},
    year={2024}
}

@misc{mistral24b,
    title={Mistral Small 3},
    author={Mistral Team},
    url={https://mistral.ai/news/mistral-small-3},
    year={2025}
}

@article{team2025gemma,
  title={Gemma 3 technical report},
  author={Team, Gemma and Kamath, Aishwarya and Ferret, Johan and Pathak, Shreya and Vieillard, Nino and Merhej, Ramona and Perrin, Sarah and Matejovicova, Tatiana and Ram{\'e}, Alexandre and Rivi{\`e}re, Morgane and others},
  journal={arXiv preprint arXiv:2503.19786},
  year={2025}
}

@misc{qwq32b,
    title = {QwQ-32B: Embracing the Power of Reinforcement Learning},
    url = {https://qwenlm.github.io/blog/qwq-32b/},
    author = {Qwen Team},
    month = {March},
    year = {2025}
}

@article{hurst2024gpt,
  title={Gpt-4o system card},
  author={Hurst, Aaron and Lerer, Adam and Goucher, Adam P and Perelman, Adam and Ramesh, Aditya and Clark, Aidan and Ostrow, AJ and Welihinda, Akila and Hayes, Alan and Radford, Alec and others},
  journal={arXiv preprint arXiv:2410.21276},
  year={2024}
}

@misc{gpt41,
  title={Introducing GPT-4.1 in the API},
  url = {https://openai.com/index/gpt-4-1/},
  author = {OpenAI},
  month = {April},
  year = {2025}
}

@article{nguyen2025sfr,
  title={SFR-DeepResearch: Towards Effective Reinforcement Learning for Autonomously Reasoning Single Agents},
  author={Nguyen, Xuan-Phi and Pandit, Shrey and Reddy, Revanth Gangi and Xu, Austin and Savarese, Silvio and Xiong, Caiming and Joty, Shafiq},
  journal={arXiv preprint arXiv:2509.06283},
  year={2025}
}

@article{jiang2025pag,
  title={PAG: Multi-Turn Reinforced LLM Self-Correction with Policy as Generative Verifier},
  author={Jiang, Yuhua and Xiong, Yuwen and Yuan, Yufeng and Xin, Chao and Xu, Wenyuan and Yue, Yu and Zhao, Qianchuan and Yan, Lin},
  journal={arXiv preprint arXiv:2506.10406},
  year={2025}
}

@article{krumdick2025no,
  title={No free labels: Limitations of llm-as-a-judge without human grounding},
  author={Krumdick, Michael and Lovering, Charles and Reddy, Varshini and Ebner, Seth and Tanner, Chris},
  journal={arXiv preprint arXiv:2503.05061},
  year={2025}
}

@article{ferrag2025llm,
  title={From llm reasoning to autonomous ai agents: A comprehensive review},
  author={Ferrag, Mohamed Amine and Tihanyi, Norbert and Debbah, Merouane},
  journal={arXiv preprint arXiv:2504.19678},
  year={2025}
}

@techreport{openai2025deepresearch,
  title       = {Deep Research System Card},
  author      = {{OpenAI}},
  institution = {OpenAI},
  type        = {Technical Report},
  month       = aug,
  year        = {2025},
  url         = {https://cdn.openai.com/deep-research-system-card.pdf},
  urldate      = {2025-08-29},
}

@article{wei2025webagent,
  title={Webagent-r1: Training web agents via end-to-end multi-turn reinforcement learning},
  author={Wei, Zhepei and Yao, Wenlin and Liu, Yao and Zhang, Weizhi and Lu, Qin and Qiu, Liang and Yu, Changlong and Xu, Puyang and Zhang, Chao and Yin, Bing and others},
  journal={arXiv preprint arXiv:2505.16421},
  year={2025}
}

@article{zhang2024aflow,
  title={Aflow: Automating agentic workflow generation},
  author={Zhang, Jiayi and Xiang, Jinyu and Yu, Zhaoyang and Teng, Fengwei and Chen, Xionghui and Chen, Jiaqi and Zhuge, Mingchen and Cheng, Xin and Hong, Sirui and Wang, Jinlin and others},
  journal={arXiv preprint arXiv:2410.10762},
  year={2024}
}

@article{ke2025mas,
  title={MAS-ZERO: Designing Multi-Agent Systems with Zero Supervision},
  author={Ke, Zixuan and Xu, Austin and Ming, Yifei and Nguyen, Xuan-Phi and Xiong, Caiming and Joty, Shafiq},
  journal={arXiv preprint arXiv:2505.14996},
  year={2025}
}

@misc{openmanus2025,
  author = {Xinbin Liang and Jinyu Xiang and Zhaoyang Yu and Jiayi Zhang and Sirui Hong and Sheng Fan and Xiao Tang},
  title = {OpenManus: An open-source framework for building general AI agents},
  year = {2025},
  publisher = {Zenodo},
  doi = {10.5281/zenodo.15186407},
  url = {https://doi.org/10.5281/zenodo.15186407},
}

@article{alzubi2025opendeepsearch,
  title={Open deep search: Democratizing search with open-source reasoning agents},
  author={Alzubi, Salaheddin and Brooks, Creston and Chiniya, Purva and Contente, Edoardo and von Gerlach, Chiara and Irwin, Lucas and Jiang, Yihan and Kaz, Arda and Nguyen, Windsor and Oh, Sewoong and others},
  journal={arXiv preprint arXiv:2503.20201},
  year={2025}
}

@article{gunjal2025rubrics,
  title={Rubrics as rewards: Reinforcement learning beyond verifiable domains},
  author={Gunjal, Anisha and Wang, Anthony and Lau, Elaine and Nath, Vaskar and Liu, Bing and Hendryx, Sean},
  journal={arXiv preprint arXiv:2507.17746},
  year={2025}
}

@article{jayalath2025compute,
  title={Compute as Teacher: Turning Inference Compute Into Reference-Free Supervision},
  author={Jayalath, Dulhan and Goel, Shashwat and Foster, Thomas and Jain, Parag and Gururangan, Suchin and Zhang, Cheng and Goyal, Anirudh and Schelten, Alan},
  journal={arXiv preprint arXiv:2509.14234},
  year={2025}
}

@article{sheng2024hybridflow,
  title   = {HybridFlow: A Flexible and Efficient RLHF Framework},
  author  = {Guangming Sheng and Chi Zhang and Zilingfeng Ye and Xibin Wu and Wang Zhang and Ru Zhang and Yanghua Peng and Haibin Lin and Chuan Wu},
  year    = {2024},
  journal = {arXiv preprint arXiv: 2409.19256}
}

@article{zhou2025variation,
    title={Variation in Verification: Understanding Verification Dynamics in Large Language Models},
    author={Zhou, Yefan and Xu, Austin and Zhou, Yilun and Singh, Janvijay and Gui, Jiang and Joty, Shafiq},
    journal={arXiv preprint arXiv:2509.17995},
    year={2025},
    url={https://arxiv.org/abs/2509.17995}
  }

@article{liu2025skywork,
  title={Skywork-Reward-V2: Scaling Preference Data Curation via Human-AI Synergy},
  author={Liu, Chris Yuhao and Zeng, Liang and Xiao, Yuzhen and He, Jujie and Liu, Jiacai and Wang, Chaojie and Yan, Rui and Shen, Wei and Zhang, Fuxiang and Xu, Jiacheng and others},
  journal={arXiv preprint arXiv:2507.01352},
  year={2025}
}

@article{ziegler2019fine,
  title={Fine-tuning language models from human preferences},
  author={Ziegler, Daniel M and Stiennon, Nisan and Wu, Jeffrey and Brown, Tom B and Radford, Alec and Amodei, Dario and Christiano, Paul and Irving, Geoffrey},
  journal={arXiv preprint arXiv:1909.08593},
  year={2019}
}

@article{dong2024rlhf,
  title={Rlhf workflow: From reward modeling to online rlhf},
  author={Dong, Hanze and Xiong, Wei and Pang, Bo and Wang, Haoxiang and Zhao, Han and Zhou, Yingbo and Jiang, Nan and Sahoo, Doyen and Xiong, Caiming and Zhang, Tong},
  journal={arXiv preprint arXiv:2405.07863},
  year={2024}
}

@article{jiang2023llm,
  title={Llm-blender: Ensembling large language models with pairwise ranking and generative fusion},
  author={Jiang, Dongfu and Ren, Xiang and Lin, Bill Yuchen},
  journal={arXiv preprint arXiv:2306.02561},
  year={2023}
}

@article{hosseini2024v,
  title={V-star: Training verifiers for self-taught reasoners},
  author={Hosseini, Arian and Yuan, Xingdi and Malkin, Nikolay and Courville, Aaron and Sordonsi, Alessandro and Agarwal, Rishabh},
  journal={arXiv preprint arXiv:2402.06457},
  year={2024}
}

@article{lou2024uncertainty,
  title={Uncertainty-aware reward model: Teaching reward models to know what is unknown},
  author={Lou, Xingzhou and Yan, Dong and Shen, Wei and Yan, Yuzi and Xie, Jian and Zhang, Junge},
  journal={arXiv preprint arXiv:2410.00847},
  year={2024}
}

@article{singh2025shelf,
  title={On the Shelf Life of Fine-Tuned LLM Judges: Future Proofing, Backward Compatibility, and Question Generalization},
  author={Singh, Janvijay and Xu, Austin and Zhou, Yilun and Zhou, Yefan and Hakkani-Tur, Dilek and Joty, Shafiq},
  journal={arXiv preprint arXiv:2509.23542},
  year={2025}
}
\bibliographystyle{iclr2026_conference}

\newpage
\appendix
\section*{Appendix}
%


\section{Extended Related Work}\label{app:related_work}
Here, we cover recent advances in automatic evaluation that are non-generative. Scalar reward models (RMs)~\citep{cobbe2021training}, which output a single scalar score for a given question $q$ and response $r$ were popularized originally within the context of reinforcement learning from human feedback (RLHF)~\citep{instructgpt_ouyang2022training,stiennon2020learning,ziegler2019fine}. Here, pairwise human preferences were used to train a RM, which is then applied during RL-based alignment, e.g., with PPO~\citep{schulman2017proximal} or DPO~\citep{hosseini2024v}. Past work has focused on dataset curation~\citep{jiang2023llm,liu2024skywork,liu2025skywork,dong2024rlhf} and experimenting with different loss formulations~\citep{liu2024skywork,lou2024uncertainty}.

The above class of reward models operate at the \textit{outcome} level, with the entire response being assigned a single score. Step-level reward models, known as process reward models, attempt to provide denser feedback by assigning each ``step'' in a response a score. Training PRMs relies on step-level labels from humans~\citep{lightman2023let} or models~\citep{duan2025efficient} or using Monte Carlo simulation to estimate step-level quality~\citep{wang2023math,luo2024improve,xiong2025stepwiser,prmlessons}. Both approaches have associated drawbacks: Step-level annotation is rarely scalable, while Monte Carlo simulation requires careful filtering to ensure high quality.

While RMs and PRMs remain popular paradigms, recent approaches have found that \textit{generative} evaluators can better leverage test-time compute for stronger evaluation performance~\citep{zhang2024generative,mahan2024generative,liu2025inference}, inspiring recent advances in generative evaluators, as discussed in~\Cref{sec:background}.

\section{Data and Training Details}\label{app:data}
\subsection{Training data}\label{app:data:data}
We enumerate our training data sources in~\Cref{tab:data}. We took efforts to decontaminate our training sets with N-gram matching approaches, following~\citet{guha2025openthoughts}. For the  \textbf{\texttt{Synthetic}} data approach, we use 12 unique generators, covering a mixture of weak and strong models: Ministral-8B~\citep{ministral8b}, Mistral-Small 24B~\citep{mistral24b}, Gemma 3 12B~\citep{team2025gemma}, Qwen2.5 7B, 32B~\citep{yang2024qwen2}, Qwen-QwQ~\citep{qwq32b}, Qwen3-30B-A3B~\citep{yang2025qwen3}, Llama-3.1-8B, Llama-3.3-70B~\citep{dubey2024llama}, GPT-4o~\citep{hurst2024gpt}, and GPT-4.1-nano, 4.1~\citep{gpt41} for seed datasets with verifiable answers. To increase diversity we randomly select a prompt template from a preset list for each question in the seed dataset and sample multiple responses at varying temperatures (e.g., 0.0, 0.3, 0.5, 0.7,...). For open-ended datasets, such as tool-use datasets, we enumerated common errors found in tool calling outputs, such as invalid input types, missing arguments, malformed json format, etc., and then programmatically injected errors into ground-truth correct responses.

\begin{table}[t!]
\caption{Data sources used to create our training set.}
\label{tab:data}
\resizebox{\textwidth}{!}{%
\begin{tabular}{lllll}
\toprule
Name &  & Task & Domain & Curation \\
& & & & Approach \\
\midrule
ActPRM & \citet{duan2025efficient} & Step-Level & Math & \textbf{\texttt{Existing}} \\
Beavertails Preference & \citet{ji2023beavertails} & Pairwise & Safety & \textbf{\texttt{Existing}} \\
Code Preference Pairs & \href{https://huggingface.co/datasets/Vezora/Code-Preference-Pairs}{Vezora/Code-Preference-Pairs} & Pairwise & Code & \textbf{\texttt{Existing}} \\
DeepScaleR & \citet{deepscaler2025} & Pairwise, Verification & Math & \textbf{\texttt{Synthetic}} \\
Folio & \citet{han2022folio} & Pairwise, Verification & NL Reasoning & \textbf{\texttt{Synthetic}} \\
FoVer & \citet{kamoi2025training} & Step-Level & Math & \textbf{\texttt{Existing}} \\
HelpSteer & \citet{wang2023helpsteer} & Single rating & Chat & \textbf{\texttt{Existing}} \\
HelpSteer2 & \citet{wang2024helpsteer} & Single rating & Chat & \textbf{\texttt{Existing}} \\
HelpSteer3 & \citet{wang2025helpsteer3,wang2025helpsteer3pref} & Pairwise & Code, NL Reasoning & \textbf{\texttt{Existing}} \\
HH-RLHF Harmless & \citet{bai2022training,ganguli2022red} & Pairwise & Safety & \textbf{\texttt{Existing}} \\
LAMP & \citet{chakrabarty2025can} & Pairwise, Single rating & Chat & \textbf{\texttt{Existing}} \\
MATH & \citet{hendrycks2021measuring} & Pairwise, Verification & Math & \textbf{\texttt{Synthetic}} \\
MemGPT & \href{https://huggingface.co/datasets/MemGPT/MemGPT-DPO-Dataset}{MemGPT/MemGPT-DPO-Dataset} & Pairwise, Verification & Tool-Use & \textbf{\texttt{Existing}} \\
OffsetBias & \citet{park2024offsetbias} & Pairwise & Chat & \textbf{\texttt{Existing}} \\
ReClor & \citet{yu2020reclor} & Pairwise, Verification & NL Reasoning & \textbf{\texttt{Synthetic}} \\
StepDPO & \citet{lai2024step} & Pairwise, Verification & Math & \textbf{\texttt{Existing}} \\
StrategyQA & \citet{geva2021did} & Pairwise, Verification & NL Reasoning & \textbf{\texttt{Synthetic}} \\
SWEGym & \citet{pan2024training} & Verification & Code & \textbf{\texttt{Existing}} \\
SWERank & \citet{reddy2025swerank} & Verification & Code & \textbf{\texttt{Existing}} \\
SynLogic & \citet{liu2025synlogic} & Pairwise, Verification & NL Reasoning & \textbf{\texttt{Synthetic}} \\
Tulu-V3-IF DPO data & \citet{lambert2024t} & Pairwise & Chat & \textbf{\texttt{Existing}} \\
WebDPO & \href{https://huggingface.co/datasets/wenhu/WebDPO}{WebDPO} & Pairwise, Verification & NL Reasoning & \textbf{\texttt{Existing}} \\
When2Call Preference Pairs & \citet{ross2025when2call} & Pairwise & Tool-Use & \textbf{\texttt{Synthetic}} \\
XLam-60K & \citet{liu2024apigen} & Pairwise, Verification & Tool-Use & \textbf{\texttt{Synthetic}} \\ 
\bottomrule
\end{tabular}%
}
\end{table}

\subsection{Training details}\label{app:data:training}
We train batch size 128 and a constant learning rate of 1e-6 and choose per-iteration rollout batch sizes of 50,000 and 250,000 for \ourmodelsmall and \ourmodellarge, respectively. In the latter case, we make the practical trade-off of RS-SFT iterations for training speed, reducing the number of times we need to reset the model for rollouts, etc. We use a modified version of the OpenRLHF framework~\citep{hu2024openrlhf} for training.

\textbf{Qwen3 cold-start SFT.} Public discussion from members of the Qwen organization indicate that post-trained versions of Qwen3 are difficult to continually finetune, and recommend starting from base models\footnote{See, for example, this \href{https://x.com/JustinLin610/status/1955077542622429321}{Twitter/X post}: ``...Instruct models after RL will pose difficulty for finetuning, but base models I don't think so...''}. Therefore, we opt to cold-start SFT with one iteration of rejection sampling data collected from Qwen2.5-32B-Instruct. As we show in~\Cref{app:additional_exp:general_purpose}, while this cold-start model does not match Qwen3-8B, \ourmodelsmall outperforms the non-thinking Qwen3-8B on many static evaluation benchmarks.

We hypothesize that a relatively short, general-purpose alignment phase prior to evaluation-specific finetuning could further improve performance. While we did not attempt this, we believe this line of experimentation is of interest for future work.

\section{Benchmark and Baseline Details}\label{app:benchmarks}
\subsection{Core benchmarks}\label{app:benchmarks:core}
For core benchmarks, we select a set of challenging and contemporary benchmarks for evaluating automatic evaluators:
\begin{itemize}
    \item JudgeBench~\citep{tan2024judgebench}: A pairwise benchmark focused on evaluating LLM-as-judge models in reasoning settings, covering math, code, logical reasoning, and knowledge-based reasoning. Responses are generated using GPT-4o.
    \item ReasoningJudgeBench~\citep{xu2025j4r}: A pairwise benchmark that covers more diverse reasoning settings, such as multi-hop, causal, and domain-specific reasoning. Responses are generated using GPT-4o.
    \item PPE Correctness~\citep{frick2024evaluate}: A pairwise benchmark that covers reasoning and instruction following tasks with objectively correct answers, using seed datasets like MATH~\cite{hendrycks2021measuring}, GPQA~\citep{rein2024gpqa}, MBPP+~\citep{evalplus}, MMLU-Pro~\citep{wang2024mmlu}, and IFEval~\citep{zhou2023instruction}. Responses are generated using a variety of weaker models, e.g., Gemma-2-9B.
    \item RM-Bench~\citep{liu2024rm}: A pairwise benchmark that evaluates how robust evaluators are to stylistic biases by evaluating on pairs of responses with subtle yet critical differences.
    \item When2Call~\citep{ross2025when2call}: A pairwise benchmark that covers appropriate selection of tools (or refusals) in response to a user prompt. We use the LLM-as-judge test split, which comprises 300 unique prompts. Each prompt has four candidate answers (refusal, direct response, tool call, follow-up question), of which one response is correct. We form all pairs, yielding 900 total pairwise comparisons.
    \item ProcessBench~\citep{zheng2024processbench}: A step-level benchmark that evaluates the ability to identify step-level errors in mathematical reasoning across easy (GSM8K and MATH) and hard (Omni-Math and OlympiadBench) questions.
    \item VerifyBench~\citep{yan2025verifybench}: A reference-based verification benchmark, comprised of Easy and Hard splits, that evaluates verifier ability to identify equivalent final answers.
\end{itemize}
For all core benchmarks, we utilize officially reported numbers when available. Otherwise, we run the corresponding baseline ourselves, using any prompt templates released with evaluators. 

For pairwise benchmarks, we select our baselines from (1) existing multi-task foundational evaluators, (2) recently released RL-trained judge models, and (3) strong-performing specialized judges:
\begin{itemize}
    \item RISE-Judge~\citep{yu2025improve}: Pairwise judges trained with SFT then DPO to perform pairwise evaluation.
    \item Self-Taught Evaluators~\citep{wang2024self}: A pairwise judge trained with iterative SFT with training data generated in the loop.
    \item EvalPlanner~\citep{saha2025learning}: Pairwise judges trained with iterative SFT and DPO on a small seed dataset, with an emphasis on learning how to plan for evaluation tasks.
    \item RM-R1~\citep{chen2025rmr1rewardmodelingreasoning}: A family of pairwise judges trained with GRPO, initialized from DeepSeek-distilled Qwen models.
    \item J1~\citep{whitehouse2025j1}: A pairwise and single-rating judge trained with GRPO.
    \item CompassJudger~\citep{cao2024compassjudger}: A family of foundational evaluators trained with large-scale SFT.
    \item Atla Selene~\citep{alexandru2025atla}: A foundational evaluator trained with large-scale preference optimization.
\end{itemize}
We run gpt-oss variants with low reasoning, as (1) \ourmodelsmall is trained initialized from gpt-oss-20B-low, and (2) evaluation often demands low-latency, making long CoT undesirable if they can be avoided. For GPT-5, we use the default API settings (medium reasoning). For pairwise benchmarks, we use reported scores from~\citet{whitehouse2025j1,xu2025j4r,liu2025inference}.

For ProcessBench, we use officially reported numbers in~\citet{zheng2024processbench}, which includes SkyworkPRMs~\citep{he_2024_16998085}, Math Shepherd PRM~\citep{wang2023math}, ActPRM~\citep{duan2025efficient}, and Qwen-Math PRMs~\citep{prmlessons}. We additionally report results from generative baselines: RL Tango~\citep{zha2025rl} and StepWiser~\citep{xiong2025stepwiser}. For VerifyBench, we use reported scores from the original paper~\citep{yan2025verifybench}.

\subsection{Downstream settings}\label{app:benchmarks:applied}
\textbf{Reward model for inference-time scaling.} We compare our models against the following baselines, representing best-in-class performers as reported in JETTS. We utilize reported numbers directly except for CompassJudger, which was not included in the original JETTS evaluation. As such, we run CompassJudger ourselves.
\begin{itemize}
    \item SFR-Judge-8B, 70B~\citep{wang2024direct}: A family of multi-task evaluators. Among the highest performing small and large judges on JETTS.
    \item Skywork-Critic-8B, 70B~\citep{skyworkcritic2024}: Two pairwise-specfic evaluators that do not output explanations. Among the highest performing small and large judges on JETTS.
    \item Self-Taught-Evaluator-70B~\citep{wang2024self}: A strong performing large judge model.
    \item CompassJudger-7B, 14B~\citep{cao2024compassjudger}: As described above.
\end{itemize}

\textbf{Verifier during GRPO training.} We adopt the settings of~\citet{ma2025general}, which train General-Reasoner, a family of reasoning LLMs of varying model sizes using the WebInstruct-Verified training dataset. In particular, we train with standard GRPO, i.e., without dynamic sampling or clip higher modifications, initializing from Qwen2.5-7B-Base. We the same conditional reward setup as General-Reasoner:
\begin{itemize}
    \item If the solution parsing fails, then reward is set to $-0.5$.
    \item If a solution successfully parsed and is deemed correct by the verifier, it is assigned a reward of $1$ plus a length penalty of:
    \begin{align*}
    -0.05 \times \min\{10, \vert\texttt{len(ground\_truth) - len(model\_response)}\vert\}.
\end{align*}
\end{itemize}
The training framework is based on the verl framework~\citep{sheng2024hybridflow}. We use rollout batch size 1024, max response length of 4096, group size of 8, a temperature of 1.0, a KL coefficient of 0.001, and a learning rate of 5e-7.

\textbf{Initialization for domain-specific finetuning.} We randomly sample 15,000 pairwise samples from AceCoder~\citep{zeng2025acecoder} and perform one round of rejection sampling. We adopt training setup of \ourmodellarge: a direct judgment ratio of 60\% and continuous curriculum. We train for 3 epochs with batch size 256 and cosine decay learning rate peaking at 1e-5. We then evaluate on CodingJudgeBench, reporting consistency accuracy. 

Note that CodingJudgeBench reports an unconventional pairwise metric, employing Z-score normalization between the two consistency runs. Their implementation is not publicly available, and their paper lacks concrete implementation details. As such, we resort to consistent accuracy, which is more commonly used in pairwise benchmarks, e.g.,~\citep{tan2024judgebench,li2023generative,xu2025does,xu2025j4r}.

\section{Ablations, analysis, and additional results}\label{sec:exp:analysis}
\begin{figure}
    \centering
    \includegraphics[width=0.8\linewidth]{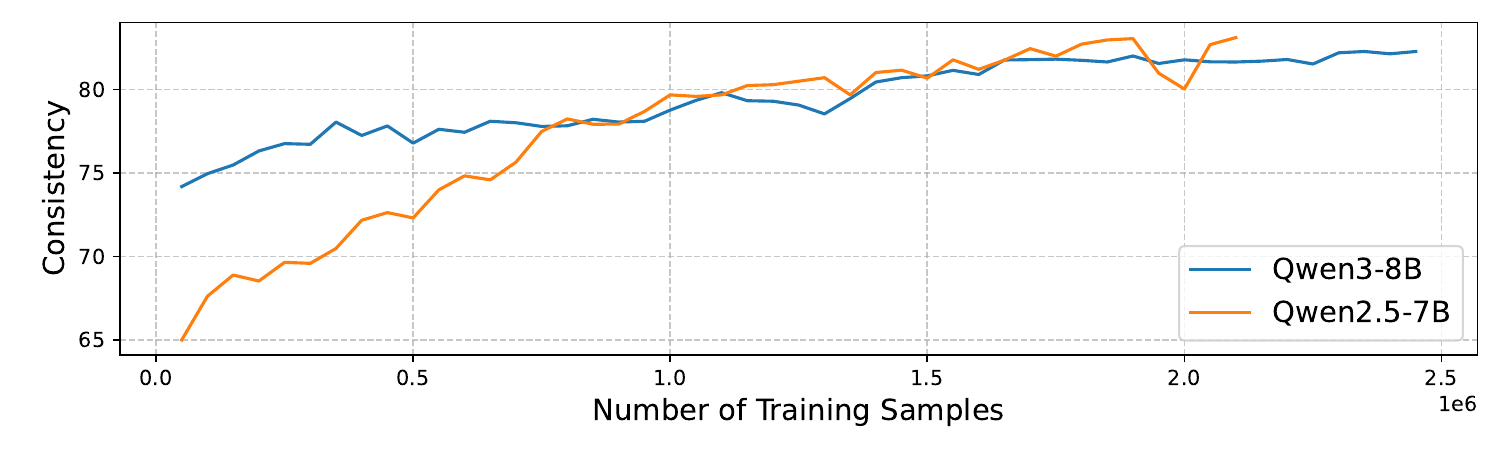}
    \caption{Pairwise positional robustness emerges as at large data scales, as shown with training \ourmodelsmall and an earlier development run initialized with Qwen2.5-7B-Instruct. \austin{Probably appendix}}
    \label{fig:consistency}
\end{figure}

\begin{table}[b!]
\caption{Ablation study varying proportion of direct judgment data, use of continuous curriculum (8B model), and ablating strategies for using direct judgment data for gpt-oss (20B model). \austin{Kinda ugly table, but should be ok. I didn't run curriculum ablation for 20b and the Cot experiment only applies to gpt-oss}}
\label{tab:ablation}
\centering
\resizebox{0.8\textwidth}{!}{%
\begin{tabular}{cccc!{\color{gray!50}\vrule}cccc}
\toprule
 \multirow{2}{*}{\begin{tabular}{@{}c@{}}{\small\% direct} \\ {\small judgment data}\end{tabular}} &  & Qwen3-8B-ColdStart Models &  &  &  & gpt-oss-20B Models &  \\
 & {\small Pairwise} & {\small ProcessBench} & {\small Average} &  & {\small Pairwise} & {\small ProcessBench} & {\small Average} \\
 \midrule 
 30 & 61.36 & 52.10 & 56.73 &  & 72.51 & 83.64 & 78.08 \\
 40 & 61.14 & \textbf{58.03} & \textbf{59.59} &  & 70.82 & 82.50 & 76.66 \\
 50 & 61.56 & 56.85 & 59.21 &  & 71.97 & 83.00 & 77.49 \\
 60 & 62.31 & 54.73 & 58.52 &  & \textbf{72.58} & 84.40 & \textbf{78.49} \\
 70 & \textbf{62.67} & 56.49 & 59.58 &  & 71.59 & \textbf{84.72} & 78.16 \\
 \midrule
 {\small Curriculum} & {\small Pairwise} & {\small ProcessBench} & {\small Average} & {\small Keep CoT? }& {\small Pairwise} & {\small ProcessBench} & {\small Average} \\
 \midrule
Yes & 61.14 & \textbf{58.03} & 59.59 & Yes & 67.80 & 81.64 & 74.72 \\
No & \textbf{61.49} & 53.43 & 57.46 & No & \textbf{69.81} & \textbf{82.55} & \textbf{76.18} \\
 \bottomrule
\end{tabular}%
}
\vspace{-1\baselineskip}
\end{table}

\subsection{Training recipe ablations.} 
In~\Cref{tab:ablation}, we ablate three components of our training recipe and report the average on our five pairwise benchmarks and ProcessBench. First, we train multiple checkpoints using RS-SFT varying the proportion of direct judgment data from 30\% to 70\%. Direct judgment data affects \ourmodelsmall and \ourmodellarge differently: endpoints (30\% or 60-70\%) show the best performance for \ourmodellarge, whereas performance peaks at 40\% for \ourmodelsmall. As such, we choose 40\% and 60\% to train \ourmodelsmall and \ourmodellarge.

We also ablate different strategies for training with direct judgment data specific to gpt-oss, which is trained to output an intermediate CoT before responding. We can either keep this CoT with direct judgment data or remove the intermediate CoT by going directly to the assistant turn. The former more closely mimics the training distribution of gpt-oss, but undermines the intended purpose of direct judgment data of isolating outcome correctness training signal, as discussed in~\Cref{sec:method:data}. The latter is out-of-training-distribution but more effectively isolates training signal. As seen in~\Cref{tab:ablation} (bottom right), removing the intermediate CoT leads to gains in both pairwise and step-level settings.

Finally, we measure the impact of the continuous curriculum as compared to a random data shuffling strategy with \ourmodelsmall. As shown in~\Cref{tab:ablation} (bottom left), the continuous curriculum leads to minimal drops in pairwise performance but large gains in ProcessBench. 

\begin{table}[t!]
\caption{Performance with and without critiques (and CoT for \ourmodellarge) for pairwise benchmarks (left) and ProcessBench (right). Directly prompting for a verdict universally improves performance for \ourmodelsmall, but degrades performance for \ourmodellarge. }
\label{tab:cot_vs_no_cot}
\resizebox{\textwidth}{!}{%
\begin{tabular}{lcccccc!{\color{gray!50}\vrule}ccccc}
\toprule
 & JudgeBench & RJB & \begin{tabular}[c]{@{}c@{}}PPE \\ Correctness\end{tabular} & \begin{tabular}[c]{@{}c@{}}RM\\ Bench\end{tabular} & When2Call & Average & GSM8K & MATH & \begin{tabular}[c]{@{}c@{}}Olympiad\\ Bench\end{tabular} & \begin{tabular}[c]{@{}c@{}}Omni\\ MATH\end{tabular} & Average \\ 
 \midrule
\ourmodelsmall & 55.71 & 51.05 & 63.8 & 79.2 & 80.33 & 66.0 & 68.5 & 67.7 & 59.9 & 58.1 & 63.5 \\
\ourmodelsmall, no critique & 60.00 & 52.60 & 64.9 & 81.8 & 86.55 & 69.2 & 68.5 & 68.6 & 59.0 & 58.5 & 63.7 \\
\ourmodellarge & 64.29 & 57.05 & 74.4 & 90.5 & 76.67 & 72.6 & 89.8 & 87.8 & 80.0 & 79.9 & 84.4 \\
\ourmodellarge, no critique or CoT & 62.00 & 55.23 & 68.9 & 85.5 & 89.11 & 72.1 & 79.8 & 73.2 & 70.0 & 70.3 & 73.3 \\ 
\bottomrule
\end{tabular}%
}
\end{table}
\subsection{How does direct judgment prompting affect performance?} 
Many settings demand low latency, such as inference-time reranking or evaluating rollouts during RL training. Here, we study how performance varies when \ourmodel are prompted to skip the critique $c$ and directly output a judgment $j$. For \ourmodellarge, this involves additionally skipping the intermediate CoT, directly outputting from the assistant turn, making out-of-distribution relative to gpt-oss-20B's original training setup. We see that performance \textit{improves} for \ourmodelsmall, but degrades for \ourmodellarge. Such results for \ourmodelsmall are in-line with prior work, which finds that direct judgment-like inference leads to minimal drops in performance. For \ourmodellarge, we hypothesize that the post-training of gpt-oss-20B instills a strong prior in favor of generating intermediate CoT. Even training with direct judgment data, removing such CoT is detrimental. Nonetheless, performance does not universally degrade, with When2Call performance increasing by nearly 13 points.  

\subsection{Robustness to pairwise positional bias emerges with data scale.} 
A known issue in pairwise evaluation is inconsistency~\citep{wang2023large}, a form of positional bias where the evaluator judgment changes based on the order of responses in the input prompt. During training, we observed that our judges become more consistent as a function of data scale;~\Cref{fig:consistency} shows the progression of pairwise consistency on the five pairwise benchmarks in~\Cref{sec:exp:core} over the course of training for \ourmodelsmall and an earlier training run which was initialized from Qwen2.5-7B-Instruct. Both models steadily become positionally robust over the course of training, with the weaker Qwen2.5-7B-Instruct showing substantial gains. This reveals that \textit{scaling evaluator training data can mitigate common judge biases}, complementing mitigation strategies that use data augmentation~\citep{saha2025learning}, label balancing~\citep{cao2024compassjudger,wang2024direct}, and RL-based reward or algorithmic methods~\citep{whitehouse2025j1,xu2025j4r}.

\begin{table}[t!]
\caption{Single rating performance, with \textbf{best} and \underline{second-best} performance in each section marked. Despite being trained with a focus on reasoning settings, \ourmodel perform competitively in single-rating evaluation in chat settings.}
\centering
\label{tab:single-rating}
\resizebox{0.5\textwidth}{!}{%
\begin{tabular}{lccc}
\toprule
 & FLASK & BiGGen Bench & Average \\ 
 \midrule
GPT-4o-mini & \textbf{0.630} & 0.600 & \underline{0.615} \\
Glider-3.8B & \underline{0.615} & 0.604 & 0.610 \\
FlowAI-Judge-3.8B & 0.400 & 0.460 & 0.430 \\
Prometheus-2-7B & 0.470 & 0.500 & 0.485 \\
Auto-J-13B & 0.350 & 0.300 & 0.325 \\
Themis-8B & 0.540 & 0.580 & 0.560 \\
SFR-Judge-8B & 0.520 & 0.590 & 0.555 \\
SFR-Judge-12B & 0.590 & 0.570 & 0.580 \\
Atla Selene 8B & 0.613 & 0.584 & 0.599 \\
LMUnit-8B & 0.600 & \textbf{0.645} & \textbf{0.623} \\
\ourmodelsmall & 0.611 & \underline{0.616} & 0.591 \\
\midrule
GPT-4o & 0.690 & 0.650 & 0.670 \\
Prometheus-8x7B & 0.540 & 0.520 & 0.530 \\
SFR-Judge-70B & \underline{0.660} & \underline{0.650} & \underline{0.655} \\
LMUnit-70B & \textbf{0.720} & \textbf{0.677} & \textbf{0.699} \\
\ourmodellarge & 0.649 & 0.616 & 0.633 \\ 
\bottomrule
\end{tabular}%
}
\end{table}
\subsection{Single-rating evaluation.} We additionally evaluate \ourmodel on Single Rating tasks with BiGGen-Bench~\citep{kim2024biggen} and FLASK~\citep{ye2023flask}, two chat-centric evaluation datasets with human annotated 1-5 ratings. We measure Pearson correlation with human annotations, and report results in~\Cref{tab:single-rating}. Single-rating is widely used evaluation task in reasoning settings, and thus constituting the smallest proportion of our training data, as shown in~\Cref{fig:dataset}. Nonetheless, \ourmodel are competitive with chat-focused judge models, with \ourmodelsmall outperforming foundational judge models like SFR-Judge-8B and 12B and \ourmodellarge approaching the performance of SFR-Judge-70B. We use reported values from baseline papers, including LMUnit~\citep{saad2024lmunit}, Atla Selene~\citep{alexandru2025atla}, and SFR-Judge~\citep{wang2024direct}.

\begin{table}[]
\caption{Comparison of \ourmodel against their initial models and other popular general-purpose models. $^\dagger$ indicates some results reported in~\citet{whitehouse2025j1} or~\citet{zheng2024processbench}.}
\label{tab:general_purpose}
\resizebox{\textwidth}{!}{%
\begin{tabular}{lccccccc}
\toprule
 & JudgeBench & ReasoningJudgeBench & PPE Correctness & RM-Bench & When2Call & Avg. consistency & ProcessBench \\
 \midrule
Qwen3-8B-ColdStart & 48.29 & 40.59 & 60.5 & 78.07 & 59.67 & 72.55 & 38.3 \\
Qwen3-8B-non-thinking & 52.27 & 43.56 & 64.8 & 79.9 & 64.78 & 74.04 & 56.7 \\
\ourmodelsmall & 55.71 & 51.05 & 63.8 & 79.2 & 80.33 & 82.28 & 63.5 \\
gpt-oss-20B (low) & 59.43 & 50.51 & 71.7 & 89.9 & 61.33 & 77.83 & 73.9 \\
\ourmodellarge & 64.29 & 57.05 & 74.4 & 90.5 & 76.67 & 82.92 & 84.4 \\
gpt-oss-120B (low) & 70.29 & 58.26 & 77.8 & 92.0 & 70.00 & 84.09 & 83.4 \\
Deepseek-R1-671B$^\dagger$ & 68.90 & 58.53 & 76.5 & 88.6 & 81.00 & - & - \\
GPT-4.1 & 66.29 & 59.68 & 78.4 & 87.8 & 64.00 & 85.54 & 77.8 \\
GPT-4o & 50.29 & 45.25 & 68.9 & 80.1 & 67.44 & 78.02 & 61.9 \\
o1-mini$^\dagger$ & 64.20 & - & 71.3 & 80.8 & - & - & 87.9\\
\bottomrule
\end{tabular}%
}
\end{table}
\subsection{Comparison against general-purpose models}\label{app:additional_exp:general_purpose}
Here, we compare \ourmodel against general-purpose LLMs, selecting popular reasoning and non-reasoning models.~\Cref{tab:general_purpose} shows our results. Our cold-start SFT for Qwen3-8B produces the weakest Qwen3 variant, as expected. However, after undergoing iterative SFT, \ourmodelsmall surpasses Qwen3-8B on multiple benchmarks, improving from 43.56 to 51.05 on ReasoningJudgeBench and 56.7 to 63.5 on ProcessBench. Likewise, we are able to improve gpt-oss-20B across the board, yielding substantial improvements in reasoning, tool-calling, and step-level evaluation. The resulting checkpoint approaches gpt-oss-120B on a number of benchmarks.

\begin{table}[t!]
\caption{Full results on JETTS. Numbers in bold indicate that the judge reranking was helpful, i.e., performance is greater than baseline (greedy) performance.}
\label{tab:jetts_full}
\resizebox{\textwidth}{!}{%
\begin{tabular}{llcccccc}
\toprule
Benchmark & Generator & Baseline & Oracle & \ourmodelsmall & \ourmodellarge & \ourmodellarge & gpt-oss-20B \\ 
& Model & Performance & Performance & & & [\texttt{CritiquePrompt}] & [\texttt{CritiquePrompt}]\\
\toprule
MATH & Llama-3.1-8B-Instruct & 24.70 & 53.47 & \textbf{35.73} & \textbf{50.83} & \textbf{49.85} & \textbf{29.83} \\
 & Llama-3.1-70B-Instruct & 43.81 & 68.35 & \textbf{53.47} & \textbf{65.41} & \textbf{64.66} & \textbf{51.06} \\
 & Qwen2.5-32B-Instruct & 57.10 & 78.17 & \textbf{65.03} & \textbf{74.32} & \textbf{74.24} & \textbf{61.56} \\
 & Qwen2.5-72B-Instruct & 62.99 & 82.78 & \textbf{70.17} & \textbf{79.98} & \textbf{78.70} & \textbf{70.32} \\ \midrule
GSM8K & Llama-3.1-8B-Instruct & 85.67 & 96.44 & \textbf{92.04} & \textbf{94.77} & \textbf{94.77} & \textbf{93.78} \\
 & Llama-3.1-70B-Instruct & 95.53 & 98.48 & \textbf{96.37} & \textbf{96.74} & \textbf{96.74} & \textbf{96.06} \\
 & Qwen2.5-32B-Instruct & 95.22 & 98.56 & \textbf{96.21} & \textbf{96.29} & \textbf{96.74} & \textbf{95.75} \\
 & Qwen2.5-72B-Instruct & 95.68 & 97.88 & \textbf{95.98} & \textbf{95.75} & \textbf{95.98} & 95.53 \\ \midrule
CHAMP & Llama-3.1-8B-Instruct & 29.26 & 60.00 & \textbf{34.07} & \textbf{44.07} & \textbf{42.22} & \textbf{35.93} \\
 & Llama-3.1-70B-Instruct & 47.41 & 71.48 & \textbf{51.85} & \textbf{58.52} & \textbf{56.67} & \textbf{55.56} \\
 & Qwen2.5-32B-Instruct & 75.19 & 85.56 & 70.00 & \textbf{77.78} & \textbf{79.26} & 74.81 \\
 & Qwen2.5-72B-Instruct & 71.48 & 85.56 & 70.00 & \textbf{73.70} & \textbf{74.81} & 67.78 \\ \midrule
MBPP & Llama-3.1-8B-Instruct & 54.50 & 76.46 & \textbf{59.79} & \textbf{68.78} & \textbf{68.25} & \textbf{63.49} \\
 & Llama-3.1-70B-Instruct & 65.08 & 83.07 & 62.17 & \textbf{67.72} & \textbf{68.78} & \textbf{67.99} \\
 & Qwen2.5-32B-Instruct & 75.40 & 84.13 & \textbf{76.72} & \textbf{80.42} & \textbf{79.37} & \textbf{79.10} \\
 & Qwen2.5-72B-Instruct & 76.19 & 84.66 & 75.40 & \textbf{78.31} & \textbf{78.31} & \textbf{78.04} \\ \midrule
HumanEval & Llama-3.1-8B-Instruct & 63.35 & 79.88 & \textbf{64.02} & \textbf{74.39} & \textbf{74.39} & \textbf{68.29} \\
 & Llama-3.1-70B-Instruct & 75.61 & 90.85 & \textbf{76.83} & \textbf{88.42} & \textbf{85.98} & \textbf{84.76} \\
 & Qwen2.5-32B-Instruct & 81.10 & 93.29 & \textbf{83.54} & \textbf{91.46} & \textbf{90.24} & \textbf{87.80} \\
 & Qwen2.5-72B-Instruct & 82.32 & 93.90 & \textbf{86.59} & \textbf{90.24} & \textbf{90.85} & \textbf{86.59} \\ \midrule
BCB & Llama-3.1-8B-Instruct & 31.67 & 56.84 & \textbf{34.82} & \textbf{41.84} & \textbf{41.23} & \textbf{39.30} \\
 & Llama-3.1-70B-Instruct & 45.44 & 62.63 & 43.86 & \textbf{46.93} & \textbf{47.54} & \textbf{45.88} \\
 & Qwen2.5-32B-Instruct & 45.53 & 65.18 & \textbf{47.02} & \textbf{49.39} & \textbf{48.95} & \textbf{48.25} \\
 & Qwen2.5-72B-Instruct & 46.67 & 60.18 & \textbf{47.54} & \textbf{49.04} & \textbf{49.30} & \textbf{48.25} \\ \bottomrule
\end{tabular}%
}
\end{table}

\subsection{Additional JETTS results.}\label{app:additional_results:jetts} In~\Cref{tab:jetts_full}, we present results full results on JETTS. Concretely, we present (a) a prompt ablation, denoted \texttt{[CritiquePrompt]}, where we prompt \ourmodellarge and gpt-oss-20B for an critique and judgment. We additionally report results for different generators than Llama-3.1-8B-Instruct. Note that unlike \ourmodellarge, gpt-oss-20B does not natively support prompting without intermediate CoT, making an even comparison with our results presented in~\Cref{fig:jetts} unfeasible. Notably, \ourmodellarge is the only judge to improve performance over greedy across all generators and benchmarks, regardless of prompt. While \ourmodelsmall is a relatively strong judge, it does not improve generator performance universally, struggling with larger generators on harder benchmarks. The trend of small evaluators struggling in helping larger generators was noted originally in JETTS. Across the board, \ourmodellarge improves in performance over gpt-oss-20B, sometimes by significant margins (e.g., 49.85 vs 29.93 for Llama-3.1-8B-Instruct MATH performance).

\section{Prompts and Examples}\label{app:prompts_examples}
\subsection{Our evaluation prompts}
Below we provide our evaluation prompts for pairwise, step-level, and verification evaluation, along with our direct judgment evaluation prompt for pairwise.

\begin{rubricbox}{Pairwise evaluation prompt for FARE}\label{prompt:pairwise}
\#\#\# System Prompt

Please act as an impartial judge and evaluate the quality of the responses provided by two AI assistants to the user prompt displayed below. You will be given assistant A's answer and assistant B's answer. Your job is to determine which assistant's answer is better.
If assistant A is better, output [A]. If assistant B is better, output [B].

Here are some rules for evaluation

(1) When evaluating the assistants' answers, identify any mistakes or inaccurate information. Focus on the content each response and select the response that is logically sound and error free.

(2) If both responses contain inaccurate information, select the response that arrives at the correct response

(3) Avoid any biases, such as order of responses, length, or stylistic elements like formatting

Before outputting your final judgment, provide an explanation of your judgment. Your explanation should discuss why your chosen response is better based on the evaluation criteria. The explanation should concretely discuss strengths and weaknesses of both answers.

After outputting your explanation, provide your final judgment. Use the following format:

Explanation: Your explanation here

Verdict: Your final verdict

\#\#\# User Prompt

[User Question]

\{question\}

[The Start of Assistant A's Answer]

\{response\_a\}

[The End of Assistant A's Answer]

[The Start of Assistant B's Answer]

\{response\_b\}

[The End of Assistant B's Answer]
\end{rubricbox}

\begin{rubricbox}{Direct judgment pairwise evaluation prompt for FARE}\label{prompt:pairwise_direct}
\#\#\# System Prompt

Please act as an impartial judge and evaluate the quality of the responses provided by two AI assistants to the user prompt displayed below. You will be given assistant A's answer and assistant B's answer. Your job is to determine which assistant's answer is better.
If assistant A is better, output [A]. If assistant B is better, output [B].

Here are some rules for evaluation

(1) When evaluating the assistants' answers, identify any mistakes or inaccurate information. Focus on the content each response and select the response that is logically sound and error free.

(2) If both responses contain inaccurate information, select the response that arrives at the correct response

(3) Avoid any biases, such as order of responses, length, or stylistic elements like formatting

Output your final judgment directly. Do not output any explanation or rationale for your decision. Use the following format:

Verdict: Your final judgment

\#\#\# User Prompt

[User Question]

\{question\}

[The Start of Assistant A's Answer]

\{response\_a\}

[The End of Assistant A's Answer]

[The Start of Assistant B's Answer]

\{response\_b\}

[The End of Assistant B's Answer]
\end{rubricbox}

\begin{rubricbox}{Step-level evaluation prompt for FARE}\label{prompt:process}
\#\#\# System Prompt

Please act as an impartial judge and evaluate the quality of the response provided by an AI assistant to the user prompt displayed below. You will be given the assistant's solution to a math problem, which is split into steps, starting with a <step [step number]> tag, where [step number] is indexed from 0. Your job is to identify which step an error occurs, if an error is present.
When evaluating the solution, consider each step separately. Evaluate the content of each step for correctness. If you encounter a mistake at <step [step number]>, output [step number] as your Verdict. If the full response is error free, then select step number -1. Avoid any biases, such as length of step, or stylistic elements like formatting.

Here are some rules for evaluation.

(1) The assistant's answer does not need to be complete or arrive at a final solution. You may receive a partially complete response. Your job is to assess the quality of each step.

(2) When evaluating the assistant's answer, identify any mistakes or inaccurate information. Focus on the content each step and determine if the step is logically valid.

(3) For each step, you should provide an explanation of your assessment. If you find an error, describe the nature and cause of the error.

(4) Avoid any biases, such as answer length, or stylistic elements like formatting.

Before providing an your final verdict, think through the judging process and output your thoughts as an explanation
After providing your explanation, you must output the corresponding step number with an error. Use the following format:

Explanation: Your explanation here

Verdict: The step number with the error or -1 if no error occurs

\#\#\# User Prompt

[User Question]

\{question\}

[The Start of Assistant's Answer]

\{response\}

[The End of Assistant's Answer]
\end{rubricbox}

\begin{rubricbox}{Reference-based verification evaluation prompt for FARE}\label{prompt:verification}
\#\#\# System Prompt

Please act as an impartial judge and evaluate if a response provided by an AI assistant (candidate answer) is consistent with a provided reference answer. 
Your job is to determine is the assistant's response is consistent with the reference answer. 

If the response is consistent, output [A].

If the response is incorrect, output [B].

Here are some rules for evaluation.

(1) Refer to the given reference answer and determine if the candidate's answer is consistent with the reference answer.

(2) The reference answer is always correct and the question is perfectly valid. Take the reference answer as the ground truth.

(3) When determining if the candidate's answer is consistent with the reference answer, only compare the final answer. Ignore any potential errors in the reasoning processes.

(4) Some answers may be expressed in different ways, such as some answers may be a mathematical expression, some answers may be a textual description, as long as the meaning expressed is the same. Before making a judgment, please understand the question and the reference answer first, and then judge whether the candidate's answer is consistent with the reference answer.

(5) Some answers may consist of multiple items, such as multiple-choice questions, multiple-select questions, fill-in-the-blank questions, etc. Regardless of the question type, the final answer will be considered correct as long as it matches the standard answer, regardless of whether the reasoning process is correct. For multiple-select questions and multiple-blank fill-in-the-blank questions, all corresponding options or blanks must be answered correctly and match the standard answer exactly to be deemed correct.

Before outputting your final judgment, provide an explanation of your judgment. Your explanation should discuss why the response is correct, incorrect, or invalid. The explanation should concretely discuss reasons for your judgment.
After outputting your explanation, provide your final judgment. Use the following format:

Explanation: Your explanation here

Verdict: Your final judgment of [A] or [B]

\#\#\# User Prompt

<|User Prompt|>

\{question\}

<|The Start of Assistant's Answer|>

\{response\}

<|The End of Assistant's Answer|>

<|The Start of Reference Answer|>

\{reference\}

<|The End of Reference Answer|>
\end{rubricbox}

\subsection{Sample evaluation rubric}\label{app:example:rubric}
Here, we provide a sample rubric that was hand-written for SWE-Rank~\citep{reddy2025swerank}. SWE-Rank data consists of contrastive pairs for training retrieval models. We re-purposed this data into a binary verification task, asking the evaluator if the retrieved code snippet was relevant for editing given a user request. ``Positive'' samples were assigned ``Correct'' labels, and ``Negative'' samples were assigned ``Incorrect'' labels.

\begin{rubricbox}{Example hand-written rubric for code retrieval samples}\label{app:rubric_example}
Here are some rules for evaluation

(1) Determine if the function provided by the assistant is a relevant candidate for editing given the user's instruction

(2) A relevant function is one means that needs to be modified in order to address the issue described in the user's instruction

(3) Modifying a relevant function does not mean is is sufficient to resolve the user's issue. That is, it is ok if modifying the function does not completely resolve the user issue, but it should make progress towards issue resolution.
\end{rubricbox}

\end{document}